\documentclass[10pt]{article} % For LaTeX2e
\usepackage[preprint]{tmlr}

\usepackage{amsmath,amsfonts,bm}

\newcommand{\figleft}{{\em (Left)}}

\newcommand{\figright}{{\em (Right)}}

\def\eqref#1{equation~\ref{#1}}
\def\1{\bm{1}}

\DeclareMathAlphabet{\mathsfit}{\encodingdefault}{\sfdefault}{m}{sl}
\SetMathAlphabet{\mathsfit}{bold}{\encodingdefault}{\sfdefault}{bx}{n}

\newcommand{\E}{\mathbb{E}}

\usepackage{hyperref}
\usepackage{url}

\usepackage{booktabs}       % professional-quality tables
\usepackage{amsfonts}       % blackboard math symbols
\usepackage{nicefrac}       % compact symbols for 1/2, etc.
\usepackage{microtype}      % microtypography
\usepackage{xcolor}         % colors

\usepackage{graphicx}
\usepackage{adjustbox}
\usepackage{enumitem}
\usepackage{wrapfig}
\usepackage{amsmath}

\usepackage{subcaption}

\usepackage[capitalize]{cleveref}
\crefname{section}{Sec.}{Secs.}
\Crefname{section}{Section}{Sections}
\crefname{table}{Tab.}{Tabs.}
\Crefname{table}{Table}{Tables}

\definecolor{todored}{HTML}{FF0000}

\renewcommand{\t}[1]{\mathrm{#1}}
\newcommand{\D}{\mathcal{D}}
\renewcommand{\L}{\mathcal{L}}
\newcommand{\xsample}{\Tilde{x}}
\newcommand{\SD}{Stable Diffusion}
\newcommand{\gen}{g}
\newcommand{\emb}{w}

\newcommand{\embcliptext}{e_\mathrm{t}}
\newcommand{\embclipimg}{e_\mathrm{i}}
\newcommand{\cliptextenc}{E_\mathrm{t}}
\newcommand{\clipimgenc}{E_\mathrm{i}}

\newcommand{\captiontitle}[1]{\textbf{#1}}

\newcommand{\ap}{AP}
\newcommand{\gap}{GAP}
\newcommand{\aplong}{Adversarial Prompts}
\newcommand{\gaplong}{Guided Adversarial Prompts}
\newcommand{\gplong}{Guided Prompts}
\newcommand{\gp}{GP}

\newcommand\quotes[1]{``{#1}''}

\definecolor{seabornblue}{HTML}{4878D0}
\definecolor{seabornviolet}{HTML}{956CB4}
\definecolor{seaborngreen}{HTML}{6ACC64}
\definecolor{seabornorange}{HTML}{EE854A}
\definecolor{seabornred}{HTML}{D65F5F}
\newcommand{\seaborngreen}[1]{\textcolor{seaborngreen}{#1}}
\newcommand{\seabornblue}[1]{\textcolor{seabornblue}{#1}}

\newcommand{\seabornorange}[1]{\textcolor{seabornorange}{#1}}
\newcommand{\seabornred}[1]{\textcolor{seabornred}{#1}}

\title{Controlled Training Data Generation with Diffusion Models}

\author{\name Teresa Yeo\footnotemark[1]\quad 
Andrei Atanov\footnotemark[1]\quad 
Harold Benoit\footnotemark[2]\quad 
Aleksandr Alekseev\footnotemark[2]\quad \\
Ruchira Ray\quad 
Pooya Esmaeil Akhoondi\quad 
Amir Zamir\\ 
\addr Swiss Federal Institute of Technology Lausanne (EPFL) \\ \\
\url{adversarial-prompts.epfl.ch}
}

\def\month{03}  % Insert correct month for camera-ready version
\def\year{2025} % Insert correct year for camera-ready version
\def\openreview{\url{https://openreview.net/forum?id=sSOxuUjE2o}} % Insert correct link to OpenReview for camera-ready version
\vspace{-5mm}
\def\projpage{\url{adversarial-prompts.epfl.ch}}

\begin{document}

\maketitle

\begin{abstract}
   We present a method to control a text-to-image generative model to produce training data useful for supervised learning.
    Unlike previous works that employ an open-loop approach via pre-defined prompts to generate new data using either a language model or human expertise, we develop an automated \textbf{closed-loop} system that involves \textbf{two feedback mechanisms}. 
    The first mechanism uses feedback from a given supervised model to find \textbf{adversarial prompts} that result in generated images that maximize the model's loss and, consequently, expose its vulnerabilities.
    While these adversarial prompts generate training examples curated for improving the given model, they are not curated for a specific target distribution of interest, which can be inefficient.
    Therefore, we introduce the second feedback mechanism that can optionally \textbf{guide} the generation process towards a desirable target distribution. 
    We call the method combining these two mechanisms Guided Adversarial Prompts.
    The proposed closed-loop system allows us to control the training data generation for a given model and target image distribution (see \cref{fig:pull}~\figright).
    We evaluate on different tasks, datasets, and architectures, with different types of distribution shifts (corruptions, spurious correlations, unseen domains) and illustrate the advantages of the proposed feedback mechanisms compared to open-loop approaches.
\end{abstract}

\section{Introduction}

The quality of data plays a crucial role in training generalizable deep learning models~\citep{taori2020measuring,miller2021accuracy,gadre_datacomp_2023}.
For a model to generalize well, its training data should be representative of the test distribution where it will be deployed.
However, real-world test conditions change over time, while training datasets are typically collected once and remain static due to high collection costs.
{This is also in contrast to evidence showing that being able to control the input data is a key contributor to how children are able to learn with only a few examples~\citep{braddick_visual_2013,lewis_multiple_2005}.}
We, therefore, focus on generating training datasets that can adapt to novel test distributions and are more sample-efficient.

Diffusion generative models~\citep{rombach2022high,ho2020denoising,sohl2015deep,nichol_glide_2022, saharia_photorealistic_2022} are trained on large-scale collections of images \citep{schuhmann_laion-5b_2022} and exhibit remarkable generalization abilities by being able to produce realistic images not seen during training. 
Additionally, unlike static datasets that they are trained on, these generative models allow us to \textit{adapt} the generation process to produce images that follow a certain conditioning. For example, they can be conditioned on textual prompts~\citep{rombach2022high} or geometric information such as depth maps~\citep{zhang2023adding}.

\begin{figure}
    \centering
    \includegraphics[width=\textwidth]{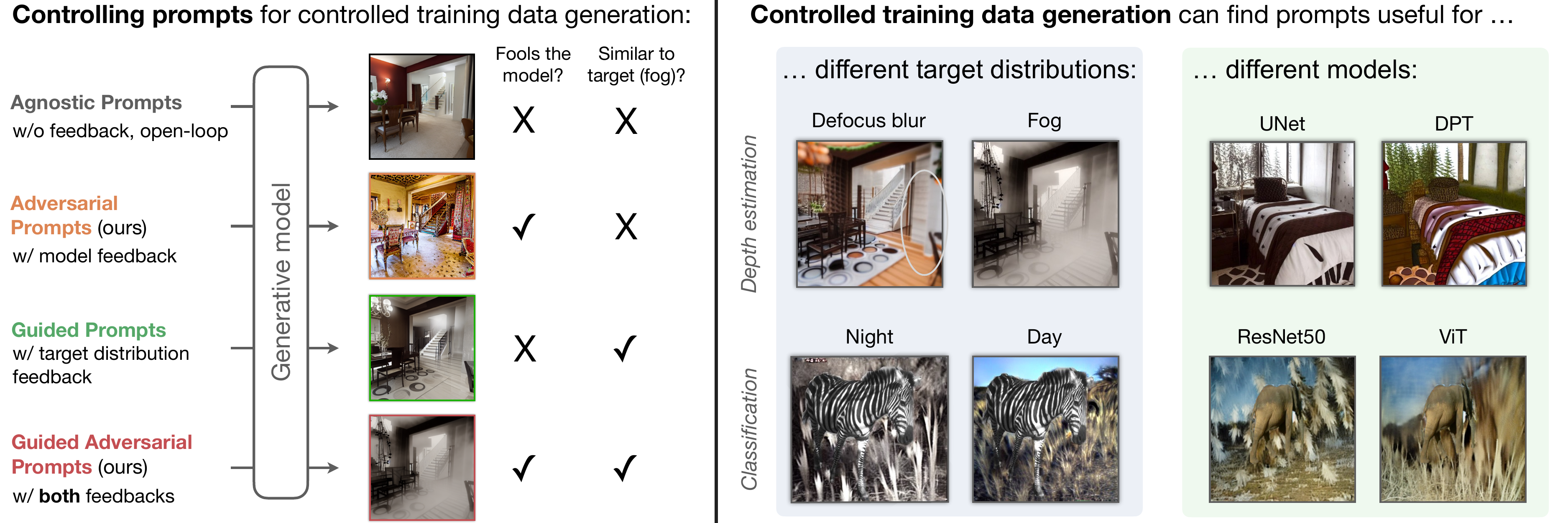}
    \caption{
    % \footnotesize
    \captiontitle{We propose a framework for the automatic generation of training examples curated for a given model and target distribution.}
    % \textbf{\textit{Left:}}
    % The proposed closed-loop system controls prompts of a conditional text-to-image generative model via two feedback mechanisms.
    % The model feedback ensures that generated examples fool a given model and the target-distribution feedback that they are visually similar to a given target image distribution.
    % \textbf{\textit{Right:}}
    % This approach allows us to find prompts and generate training examples useful for a specific target distribution and a supervised model
    % (the oval highlights the effect of defocus blur in the corresponding example.)
    % While the differences in generations for different models might appear unintuitive, they, indeed, result in useful model-specific data (see \cref{sec:analysis}).
    \textbf{\textit{Left:}}
    The method controls the prompts of a conditional text-to-image generative model via two feedback mechanisms. The `model feedback' ensures that generated examples expose vulnerabilities by fooling a given model, resulting in \seabornorange{Adversarial Prompts}. The `target distribution feedback' guides the generations towards a desirable target distribution (fog in this example.) Combining both feedback mechanisms results in \seabornred{Guided Adversarial Prompts}, leading to generations that both fool the model and fit the desirable target distribution. 
    \textbf{\textit{Right:}}
    This approach allows us to find prompts that generate training examples useful for different target distributions, supervised models, and tasks
    (the oval highlights the effect of defocus blur in the corresponding example.)
    While the differences in generations for different models might appear subtle or unintuitive, they, indeed, result in more useful training data (see \cref{sec:analysis}).
    }
    \label{fig:pull}
\end{figure}

Recent works explore the use of diffusion models to generate training data for supervised learning with promising results \citep{sariyildiz2023fake,dunlap2023diversify,he2022synthetic}.
They guide the generation process using text prompts to accomplish two goals: 1) produce aligned image-label pairs for supervised training and 2) adapt the generated images to a certain target distribution.
However, these methods find conditioning text prompts in an open-loop way, i.e., uninformed of the specific training scenario in hand, by either using a language model~\citep{dunlap2023diversify} or heuristics~\citep{sariyildiz2023fake}.
In other words, they \textit{lack an automatic feedback mechanism} that can refine the found text prompts to produce more curated and useful training data. 
% \aatodo{Furthermore, it has been argued that being able to control the input data is a key contributor to how children are able to learn with few examples~\citep{braddick_visual_2013,lewis_multiple_2005}.}

In this work, we propose two feedback mechanisms to find prompts for generating useful training data.
The first mechanism finds prompts that result in generations that maximize the loss of a particular supervised model, thus, \textit{reflecting its failure modes}.
We call them \aplong~(\ap).
This mechanism ensures that we find not only novel prompts, which may still produce images that the model already performs well on, but adversarial prompts that produce images with high loss, and, thus, useful for improving the model~\citep{deng_adversarial_2021} (see exemplar generations for \ap~in \cref{fig:pull,fig:wb-results,fig:depth-gen-comparison})

One given model can perform poorly on \textit{multiple} distribution shifts, and hoping to cover all of the shifts via Adversarial Prompts is an inefficient strategy (e.g., see the difference between the \ap~generation and the target distribution, fog, in \cref{fig:pull}~\figleft).
Therefore, we introduce an additional \textit{target distribution-informed} feedback mechanism that can find prompts that generate images fitting a desirable target distribution we want to adapt to.
To implement it, we assume access to a textual description of the target distribution and/or \textit{a few unlabeled} images from it.
We then optimize a similarity metric between CLIP~\citep{radford_learning_2021} embeddings of the generated examples and the target description.
We call prompts that combine both mechanisms \gaplong~(GAP).
Compare the \aplong~and \gaplong~in \cref{fig:pull,fig:wb-results,fig:iwild-joint,fig:depth-gen-comparison} to see the effect of the guidance feedback in steering the generations towards a specific target distribution.

We perform evaluations on different tasks (image classification, depth estimation), datasets with distribution shifts (Waterbirds~\citep{sagawa2019distributionally}, iWildCam~\citep{beery2021iwildcam,koh2021wilds}, Common Corruptions~\citep{hendrycks2019benchmarking}, 3D Common Corruptions~\citep{kar_3d_2022}), and architectures (convolutional and transformer) and show supportive results. 
\section{Related Work}
\label{sec:related_work}

\textbf{Open-loop data generation} methods use pre-defined controls to guide the generative process and produce novel training examples.
One line of work uses GANs~\citep{jahanian2021generative, chai_ensembling_2021, ravuri_classification_2019} and pre-define a perturbation in their latent space to generate novel examples.
More recent works adopt text-to-image diffusion models and use pre-defined prompt templates~\citep{sariyildiz2023fake,yuan2022not,he2022synthetic} or use a language model to generative variations of a given prompt~\citep{yuan2022not}.
These methods require \textit{anticipating the kind of data that will be seen at test-time} when defining the prompts. 
On the other hand, our CLIP guidance mechanism allows us to generate images similar to the target distribution.
\citet{dunlap2023diversify} also approach this problem by using a captioning and language model to summarize a target distribution shift into a text prompt. However, this summarization process is not informed of the generations, and, thus, does not guarantee that the text prompt will guide the generation process to images related to the target distribution. 
Finally, these methods are not model-informed and do not necessarily generate images \textit{useful} for training a given model.

\textbf{Closed-loop data generation} methods guide the generation process via an automatic feedback mechanism.
They control the latent space of GANs~\citep{besnier2020dataset} or VAEs~\citep{wong2020learning} models, NeRF~\citep{dong2022viewfool}, or the space of hand-crafted augmentations~\citep{cubuk2018autoaugment} to generate data that maximizes the loss of the network on the generated data.
Similarly, \citet{jain2022distilling} uses an SVM to identify the failure modes of a given model and uses this information to generate training data with a diffusion model.
Our method employs a similar adversarial formulation (in conjunction with target distribution guidance) but performs the optimization in the text prompt space of recently developed diffusion models.

\textbf{``Shallow'' data augmentation} techniques apply simple hand-crafted transformations to training images to increase data diversity and improve the model's generalization.
Examples of such transformations are color jitter, random crop, and flipping, etc.
To produce more diverse augmentations, methods like RandAugment~\citep{cubuk2020randaugment} and AugMix~\citep{hendrycks2019augmix} combine multiple of such simple transformations, and Mixup~\citep{zhang2017mixup} and CutMix~\citep{yun2019cutmix} methods use transformations that can combine multiple images.
AutoAugment~\citep{cubuk2018autoaugment} and adversarial training~\citep{madry2017towards} build a closed system to tune the parameters of the applied augmentations but are inherently limited by the expressiveness of the simple transformations. 
In contrast, our method uses expressive diffusion models, which results in images that are more diverse and realistic than those produced by ``shallow'' augmentations. 

\textbf{Controlling diffusion models.} Methods like ControlNet~\citep{zhang2023adding} and T2I-Adapter~\citep{mou2023t2i} adapt a pre-trained diffusion model to allow for additional conditioning e.g., edge, segmentation, and depth maps. We employ these models for generation as it allows us to generate paired data for different tasks, given the labels from an existing dataset. 
Editing methods aim to modify a given image, either via the prompt~\citep{hertz2022prompt}, masks~\citep{couairon2022diffedit}, instructions~\citep{brooks2023instructpix2pix} or inversion of the latent space~\citep{mokady2023null,huberman2023edit}. In contrast, personalization methods aim to adapt diffusion models to a given concept e.g., an object, individual, or style. Popular examples include textual inversion~\citep{gal2022image} and DreamBooth~\citep{ruiz2023dreambooth}, which aim to find a token to represent a concept given several images of that concept. The former freezes the diffusion model, while the latter fine-tunes it. Extensions of these works learn to represent multiple concepts \citep{avrahami2023break,han2023svdiff}.
In our work, we adopt an approach similar to textual inversion to steer the diffusion model, but our method can also be used with other controlling mechanisms.

\section{Method}
\label{sec:method}

\begin{figure}[t]
    \centering    
    \vspace{-8mm}
    \includegraphics[width=0.95\textwidth]{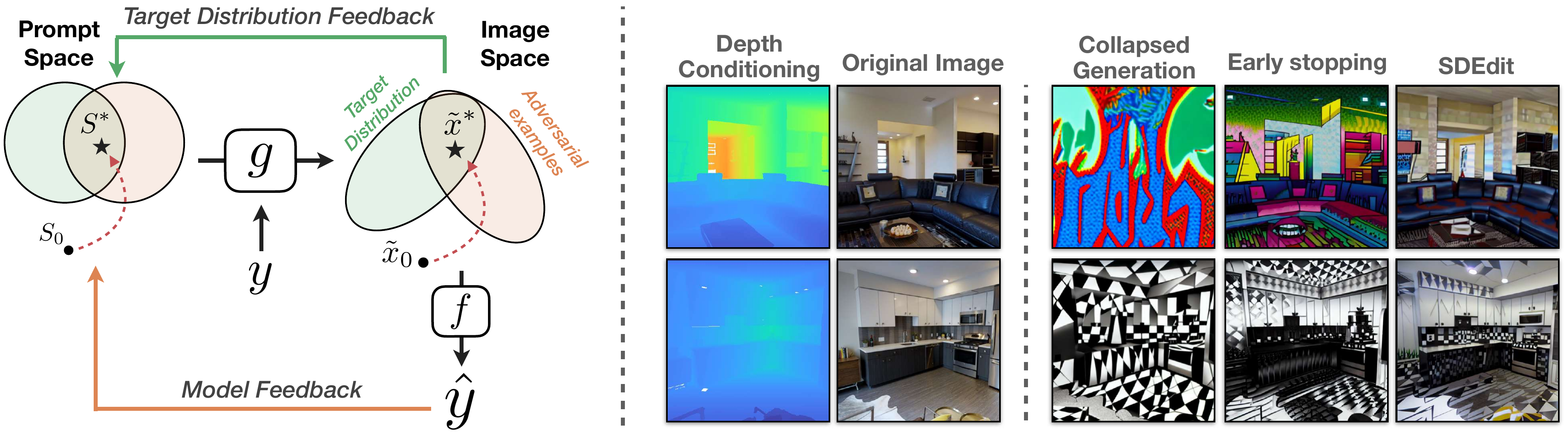}
    \caption{
    % \footnotesize 
    \textbf{Left: An overview of how we generate training data} for a given supervised model $f$ and target distribution. Suppose $g$ is generative model that generates images conditioned on a text prompt, $S$ and label, $y$. 
    A supervised model trained to perform a task, e.g., image classification or depth estimation, is denoted by $f$. 
    We aim to find prompts that would generate training data useful for $f$, and we do so via two feedback mechanisms.
    The first mechanism makes use of feedback from $f$ to get \seabornorange{\aplong}.
    % In particular, we maximize the loss on the predictions from $f$ to get \seabornorange{\aplong}. 
    % The space of \seabornorange{\aplong} and corresponding adversarial examples~(orange circle).
    Since there can be many adversarial examples not relevant to a particular target distribution, we introduce a second feedback mechanism that guides the prompts towards a certain target distribution.
    % ~(green circle).
    Combining the two mechanisms results in \seabornred{\gaplong}.
    % , represented by $S^*$. 
    This results in generations that are both \textit{relevant to the target distribution} and where $f$ \textit{does not perform well}.
    {\textbf{Right: Ways to alleviate the misalignment of the generation with its conditioning.} The third column onwards shows some examples of generations from depth maps that \textit{do not follow} the depth conditioning. See the first and second columns for the original image and its depth label. There are several ways to constrain the generation to alleviate this misalignment.
    \textbf{1.} Early stopping involves stopping the adversarial optimization when the loss reaches a certain threshold. The resulting generations from early stopping are shown in the fourth column. 
    \textbf{2.} SDEdit~\citep{meng_sdedit_2021} involves conditioning the generation process on the original image. This mechanism is applied during generation with the adversarial prompts i.e., applying SDEdit to the prompts that generated the images in the third column results in the last column generations. Both SDEdit and early stopping are able to improve the alignment of the generations with depth conditioning.}}\vspace{-2mm}
    \label{fig:method_collapse}
\end{figure}

We begin this section by formalizing our problem setting and describing how diffusion models can be used to generate training data (\cref{sec:prelim}).
We then introduce two feedback mechanisms to find prompts that are informed of the failure modes of a given model (\cref{sec:adv-prompt}) and relevant to a given target distribution (\cref{sec:clip-guidance}). 

\subsection{Preliminaries}
\label{sec:prelim}

\textbf{Problem Formulation.}
We consider the supervised learning problem, where a model $f:\mathcal{X} \to \mathcal{Y}$ learns a mapping from the image space $\mathcal{X}$, to a target space $\mathcal{Y}$, e.g., depth estimation or semantic classification.
The model $f$ is trained using a training dataset $\D_\text{train}$ and tested on a new set $\D_\text{test}$ that exhibits a distribution shift w.r.t. the training data, e.g., corruptions~\cite{hendrycks2019benchmarking}.
Our goal is to generate additional \textit{synthetic} training data $\D_\text{syn}$ to adapt the model and improve its performance under the distribution shift.
% To apply the target-informed feedback mechanism described in~\cref{sec:clip-guidance}, we assume access to some information about the test distribution, either from text descriptions or a few samples of \textit{unlabeled} images.

\textbf{Text-to-image Diffusion Models.}
We use the \SD~\citep{rombach2022high} text-to-image diffusion model as the basis for our generator $\gen$.
Given a textual prompt $c$, \SD~is capable of synthesizing realistic images following the textual conditioning.
However, in general, for a given task, e.g., depth estimation, a textual prompt alone may not be sufficient for controlling the generation well enough to produce aligned image-label examples.

\textbf{Generating aligned training examples.}
We employ the following two approaches to condition the generative model $\gen$ on a label $y$ from the given training datasets and sample aligned training examples $(\xsample, y)$.
For the depth estimation task, we use the ControlNet~\citep{zhang2023adding} model which extends the conditioning mechanisms of the \SD~to accept various spatial modalities, e.g., depth maps and segmentation masks.
Specifically, we use ControlNet v1.0 with depth conditioning\footnote{\url{https://github.com/lllyasviel/ControlNet}}.
For semantic classification tasks, we utilize the foreground object masks and use an in-painting technique proposed by \citet{lugmayr_repaint_2022} that preserves the masked region throughout the denoising process, essentially keeping it intact.
These mechanisms provide us with a generative model conditioned both on a text prompt $c$ and label $y$.
We denote the resulting distribution modeled by this generative model as $\gen(y, c)$.

\subsection{Model-Informed Generation with Adversarial Prompt Optimization}
\label{sec:adv-prompt}
Our first feedback mechanism aims at generating training examples that reflect the failure modes of a given model $f$.
An automatic way to do so is via \textbf{adversarial optimization}, which finds the \quotes{worst case} failure modes of $f$. 
More precisely, we find a prompt $c$ that generates images $\xsample\sim g(y, c)$ that maximize the supervised loss $\L(f(\xsample), y)$, e.g., $l_1$ loss for depth estimation.
Since the usual prompt space is discrete (text tokens) and challenging to optimize over, we employ the approach introduced by~\citet{gal2022image} and, instead, optimize over the corresponding continuous embedding space. For ease of notation, \quotes{prompt space} will implicitly refer to the continuous embedding space instead of the discrete token space. We construct a prompt $c_\emb$ out of $n$ new ``placeholder'' tokens, i.e., $c_\emb = (c_{\emb_1}, \dots, c_{\emb_n})$, and find their corresponding embedding weights $\{\emb_i\}_{i=1}^n$ by solving the following optimization problem:
\begin{equation}
\label{eq:adversarial-opt}
    \emb_\mathrm{AP} = \arg\min_\emb \E_y~\E_{\xsample \sim \gen(y, c_\emb)} \L_\mathrm{adv}(f(\xsample), y),
\end{equation}
where $\L_\mathrm{adv} = -\L$ and $y$ is sampled from $\D_\t{train}$.
Note that the sample $\xsample$ is differentiable w.r.t. the embeddings $w$ which allows us to use gradient-based optimization. We call the prompts that result from solving the above optimization problem \aplong~(\ap).

% \begin{wrapfigure}{r}{0.5\textwidth}
% \vspace{-3mm}
% \centering 
% \includegraphics[width=0.48\columnwidth]{figures/collapsed-gens_compressed.pdf}
% \caption{\footnotesize{\textbf{Ways to alleviate the misalignment of the generation with its conditioning.} The third column onwards shows some examples of generations from depth maps that \textit{do not follow} the depth conditioning. See the first and second columns for the original image and its depth label. There are several ways to constrain the generation to alleviate this misalignment.
% \textbf{1.} Early stopping involves stopping the adversarial optimization when the loss reaches a certain threshold. The resulting generations from early stopping are shown in the fourth column. 
% \textbf{2.} SDEdit~\citep{meng_sdedit_2021} involves conditioning the generation process on the original image. This mechanism is applied during generation with the adversarial prompts i.e., applying SDEdit to the prompts that generated the images in the third column results in the last column generations. Both SDEdit and early stopping are able to improve the alignment of the generations with depth conditioning.}}
% \label{fig:failure-cases}  \vspace{-5mm}
% \end{wrapfigure}

\textbf{Avoiding $(\xsample, y)$ alignment collapse.}
\label{sec:alignment-collapse}
While the adversarial objective in \cref{eq:adversarial-opt} aims to fool the model $f$, it may instead fool the label-conditioning mechanism of the generative model $\gen$, resulting in $c_{\emb_\t{adv}}$  generating samples $\xsample\sim\gen(y, c_{\emb_\t{adv}})$ that are not faithful to $y$ (see \cref{fig:method_collapse}~\figright).
To avoid this collapse, we use the following techniques.
% To avoid this, we further constrain the expressiveness of the generation process. There are several ways to do so.

First, we limit the expressivity of the generative model using the SDEdit method~\citep{meng2021sdedit}.
It conditions the generation process on the original image $x$ by starting the denoising process from its noised version instead of pure noise, constraining the expressive power of the generative model and producing samples closer to the original image $x$, {and, therefore, more faithful to $y$.}

Additionally, we implement constraints w.r.t. $\L_\mathrm{adv}$.
For depth estimation, we employ an early stopping criterion and stop the adversarial optimization when the loss reaches a certain threshold. {We also experimented with constraining the embedding weights $w$ to be close to existing vectors from the vocabulary in E.q.~\ref{eq:adversarial-opt}. However, we found that this significantly reduces the expressivity of the generations. Thus, we do not constrain $w$ during optimization.}
For image classification, we found that adversarial optimization of the standard cross-entropy loss can lead to the generation of images of another class (e.g., generating elephant, when $y=\mathrm{giraffe}$).
We, therefore, maximize the entropy of the model's predictions, i.e., its uncertainty, instead, which we found to lead to the generation of aligned $(\xsample,y)$ examples.
Please see \cref{app:iwild-alignment-collapse} for more details and examples.

% For semantic classification, choosing  $\L_\mathrm{adv}$ to be the negative cross-entropy loss, although natural, may not be a good choice. Indeed, for iWildCam, although we keep the class mask intact, we observed that optimizing the negative cross-entropy loss may lead to the generation of another class somewhere else in the image,  e.g. an elephant is generated next to a giraffe, destroying the $(\xsample,y)$ alignment. Thus, for iWildCam, we choose to maximize the uncertainty or entropy of the model's prediction on the generated images. We provide more details in the~\cref{app:iwild-alignment-collapse}.

% Finally, our CLIP~\citep{radford_learning_2021} guidance loss introduced in \cref{sec:clip-guidance} further constrains possible perturbations to a target distribution and helps to avoid the generation of non-realistic images.

\subsection{Target Distribution Informed Generation}
\label{sec:clip-guidance}
The adversarial formulation above finds prompts that reflect the failure modes of $f$. 
Without any information about the target distribution, improving the model on the worst-performing distributions is one of the best strategies, which, indeed, improves performance in multiple cases (see \cref{fig:iwild-joint} and \cref{tab:depth-results}).
However, a given model typically has many failure modes and many possible distribution shifts that can occur at test time.
Adapting to all of them using only the adversarial feedback mechanism might be inefficient if the goal is to adapt to a specific target distribution instead of improving average performance.
Thus, we introduce the second feedback mechanism to inform the prompt optimization process of the target image distribution.
It only requires access to simple text descriptions (e.g., `fog' to adapt to foggy images) or a small number~($\sim100$) of \textit{unlabelled} images.

We implement the target-informed feedback mechanism using CLIP~\citep{radford_learning_2021} guidance.
Specifically, we assume access to either textual descriptions of the target image distribution $\{t_j\}$, a few unlabeled image samples $\{x_j\}$, or both.
We construct the corresponding text and image guidance embeddings as $\embcliptext = \t{avg}(\{\cliptextenc(t_j)\})$ and $\embclipimg = \t{avg}(\{\clipimgenc(x_j)\})$, where $\cliptextenc$ and $\clipimgenc$ denote, respectively, the CLIP text and image encoders, and $\t{avg}$ stand for averaging.
We then use the following guidance loss:
\begin{equation}
    \label{eq:clip-guidance}
    \L_\t{G}(\xsample, c_\emb) = \lambda_\t{t} \L_\t{t}(\cliptextenc(c_\emb), \embcliptext) + \lambda_\t{i} \L_\t{i}(\clipimgenc(\xsample), \embclipimg),
\end{equation}
where we take $\L_\t{t}$ to be $l_2$ norm between two embeddings and $\L_\t{i}$ to be the negative cosine similarity, as we found it to perform the best. See the~\cref{app:classification-additional-analysis} for the results of this ablation.
Note that our formulation allows the use of only text or only images for guidance (setting corresponding $\lambda$s to 0) based on the available information.
In \cref{sec:ti-guidance}, we also show that one can construct a similarly effective guidance mechanism without relying on the CLIP model.
% based on the available information, one can also use only one of the two guidance losses.

Finally, we combine both adversarial, \cref{eq:adversarial-opt}, and guidance, \cref{eq:clip-guidance}, losses to form the final objective:
\begin{equation}
\label{eq:adversarial-opt_and_clip-guidance}
    \emb_\mathrm{GAP} = \arg\min_\emb \E_y~\E_{\xsample \sim \gen(y, c_\emb)}  \left[ \L_\mathrm{adv}(f(\xsample), y) + \L_\t{G}(\xsample, c_\emb) \right].
\end{equation}
We call the prompts that result from solving \cref{eq:adversarial-opt_and_clip-guidance}, \gaplong~(\gap). See the~\cref{app:waterbirds-implementation-dets,app:iwildcam-implementation-dets} for further implementation details.

\section{Experiments}
\label{seq:experiments}
We perform experiments in three settings: domain generalization via camera trap animal classification on the iWildCam~\citep{beery2021iwildcam} dataset, bird classification with spurious correlation with the Waterbirds~\citep{sagawa2019distributionally} dataset, and depth estimation with the Taskonomy dataset~\citep{zamir_taskonomy_2018, zamir_robust_2020}.
For depth estimation, we evaluate on distribution shifts from Common
Corruptions~\citep{hendrycks2019benchmarking} (CC), 3D Common Corruptions~\citep{kar_3d_2022} (3DCC) applied on the Taskonomy~\citep{zamir_taskonomy_2018} test set and cross dataset shift from the Replica~\citep{straub_replica_2019} dataset.

\subsection{Semantic Image Classification}
\label{sec:classification}
\begin{figure*}[t]
    \centering
    \vspace{-6mm}
    \includegraphics[width=0.95\textwidth]{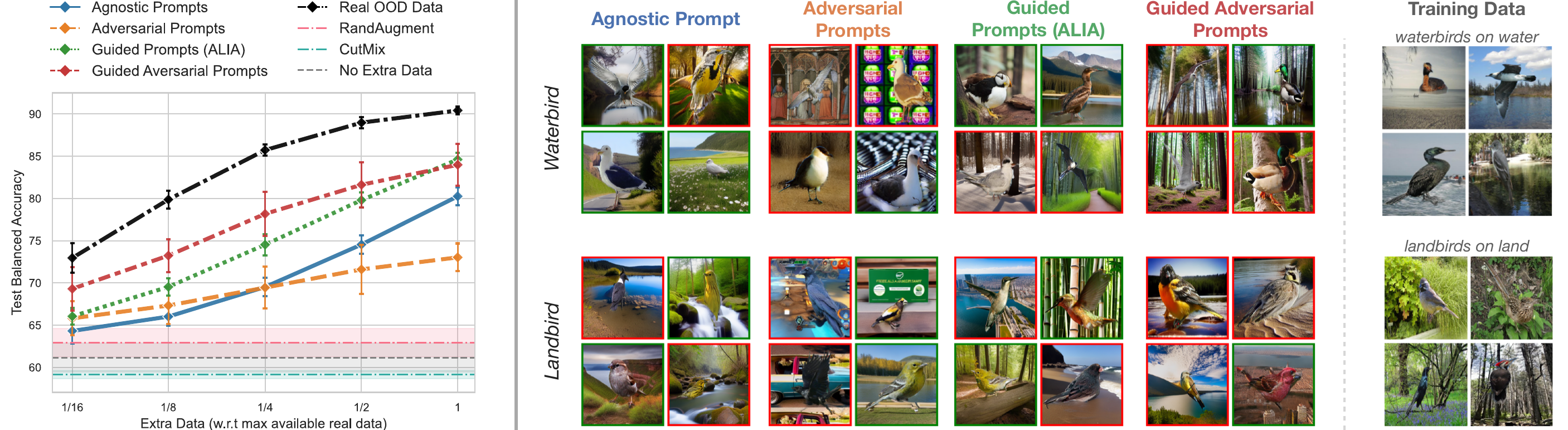}
    \caption{
    % \footnotesize
    \captiontitle{\gaplong~generate counterfactual examples not present in the original training dataset and achieve better data efficiency than other prompts.}
    \textbf{\textit{Left:}} 
    Test performance on a balanced test set (waterbirds and landbirds appear on both land and water) of a model trained on the original spuriously correlated dataset, while varying the number of extra data points generated by different types of prompts {(see~\cref{app:add-gen-data} for results on 2X data)}.
    We run each experiment with three seeds and report the mean and standard deviation.
    \textbf{\textit{Right:}} Exemplary generations for each prompt type and original training examples (where background is correlated with the bird type.)
    % A red frame signifies that the model trained only on the original data (No Extra Data) misclassifies the image, and a green frame stands for the correct prediction.
    Frames show whether the model trained without extra data made a correct prediction (green) or not (red) on the corresponding image.
    % We find that having a guidance mechanism towards the target image distribution consistently improves on top of the \seabornblue{Agnostic Prompts} baseline (``nature'' prompt).
    % \seabornorange{\aplong}, while fooling the model, generates images that are different from the target distribution and, thus, not useful to adapt the model to it.
    As expected, the adversarial mechanism produces examples that fool the model, and the guidance mechanism steers the generation toward the target distribution.
    Combining both mechanisms in \seabornred{\gap}~leads to the \textbf{generation of counterfactual examples} (waterbirds on land and landbirds on water), missing in the training dataset, and improved data efficiency compared to only target-informed \seaborngreen{\gp}.
    % Unlike \seaborngreen{\gplong} that uses the same prompts to generate images for both classes, \seabornred{\gap}~finds prompts that generate images the model fails on, this leads to generation of waterbirds on land and landbirds on water, the combinations not present in the original training data, which are necessary data samples for the model to learn the bird predictive feature.
    %This demonstrates that \seabornred{\gap}~can find more useful examples for training the model.
    }
    % \vspace{-3mm}
    \label{fig:wb-results}
\end{figure*}

\textbf{Waterbirds}~\citep{sagawa2019distributionally} is a dataset constructed by pasting an image of a waterbird or landbird from the CUB~\citep{wah_caltech-ucsd_2011} dataset, which represents the label $y$, onto a background image of ``land'' or ``water'' from the Places~\citep{zhou_learning_2014} dataset.
We follow \citet{dunlap2023diversify} and use only images of waterbirds on water and landbirds on land as $\D_\t{train}$, i.e., the background is completely correlated with the label in training data (see \cref{fig:wb-results}-right).
The test set $\D_\t{test}$ contains all four combinations of bird types and backgrounds (see \cref{fig:waterbirds-dataset}).

\textbf{iWildCam}~\citep{beery2021iwildcam, koh2021wilds} is a domain generalization dataset made up of images captured from camera traps placed in various locations around the world. The goal is to generalize to photos taken from new camera deployments. We follow~\citet{dunlap2023diversify} and create a 7-way classification task (background, cattle, elephant, impala, zebra, giraffe, dik-dik), use two locations that are not present in the training or validation as $\D_\t{test}$, and fix the number of additional generated images for finetuning to 2224. 

For both datasets, we use the ResNet50~\citep{he2016deep} model. We compare the following methods for generating additional synthetic data $\D_\t{syn}$ (we provide more details in~\cref{app:classification}):
\begin{itemize}[leftmargin=*,label={}] \vspace{-2mm}%[leftmargin=2.0mm,label={}]
    \setlength{\itemsep}{0.4pt}%
    \setlength{\parskip}{2pt}%
	\item \textit{No Extra Data}: Only $\D_\t{train}$ is used, without generating extra data.
	\item \textit{Augmentation baselines}: We use two data augmentation methods commonly used in recent literature: CutMix~\citep{yun2019cutmix} and RandAugment~\citep{cubuk2020randaugment}. These baselines do not use generative models.
	\item \textit{Agnostic Prompts}: We use a prompt that is uninformed of both the model and the specific target distribution. Similar to ALIA~\citep{dunlap2023diversify}, we use a prompt \textit{``nature''} for Waterbirds and the prompt template \quotes{\textit{a camera trap photo of \{class name\}}} for iWildCam.
	\item \textit{\gplong}: We compare to ALIA~\citep{dunlap2023diversify}, which uses captioning and language models to summarize a target distribution shift into text prompts.
    This results in seven prompts for Waterbirds and four prompts for iWildCam. See~\cref{sec:alia} for more details, specific prompts, and a discussion on the differences between ALIA and our method.
    \item \textit{\aplong}: We use the model previously trained on $\D_\t{train}$ (i.e., no extra data) as the target model $f$ and find adversarial prompts following \cref{eq:adversarial-opt}. We find four prompts per class for Waterbirds, eight in total, and four prompts in total applied to all classes for iWildCam.
    \item \textit{\gaplong}: We use the same setting as in \aplong~and apply additional CLIP guidance to adapt to a target distribution shift, following \cref{eq:adversarial-opt_and_clip-guidance}. For Waterbirds, we apply text guidance using ALIA prompts as the textual description of the target distribution. For iWildCam, we use image guidance and partition the target test distribution into four groups based on two attributes that have a significant impact on the visual characteristics of the data: the test location (first or second) and time of the day (day or night). We sample 64 \textit{unlabelled} images randomly from each group. We optimize for one guided adversarial prompt per group, resulting in four prompts in total.
\end{itemize} 
% \vspace{-2mm}

\textbf{Training details.} 
For supervised training, we follow \citet{dunlap2023diversify} and use the ResNet50 model for $f$.
In~\cref{sec:analysis}, we also explore how adversarial prompts change with the model by using ViT~\citep{dosovitskiy2020image}.
For prompt optimization, we use Adam~\citep{kingma_adam_2017} with \texttt{1e-3} learning rate.
For efficiency, we use five denoising steps during optimization and 15 for the final generations, both using the DDIM~\citep{song2020denoising} scheduler.
As $\L_\t{adv}$, we use prediction's entropy for iWildCam to avoid alignment collapse between $\xsample$ and $y$ as discussed in \cref{sec:alignment-collapse}, and negative cross-entropy for Waterbirds where we did not observe collapse.
Please see \cref{app:classification} for more details on the experimental setup.

\textbf{Waterbirds results.}
\cref{fig:wb-results}~\figleft~shows the performance of each method, and \cref{fig:wb-results}~\figright~demonstrates the corresponding generated examples for each method.
% We can observe the following trends.
First, we find that while the performance of \aplong~is on par with Agnostic Prompts in a low-data regime, it performs worse with more generated data.
Looking at the corresponding generated images, we find that while being adversarial, they appear different from images from the target distribution, which can explain the inferior performance in this particular case.
Second, using \gplong{} informed of the target distribution leads to a consistent improvement over Agnostic Prompts, and the corresponding images look more similar to the real images from the target distribution.
Finally, combining both feedback mechanisms in \gaplong{} results in more data-efficient generations, outperforming all other methods in the low-data regime.
Looking at the corresponding generations, we find that \gap~tends to generate only combinations of the bird type and background that are missing in the training dataset, i.e., waterbirds on land and landbirds on water, which can explain its higher sample efficiency.

\begin{figure}[t]
    \centering 
    \vspace{-10mm}
    \includegraphics[width=0.95\columnwidth]{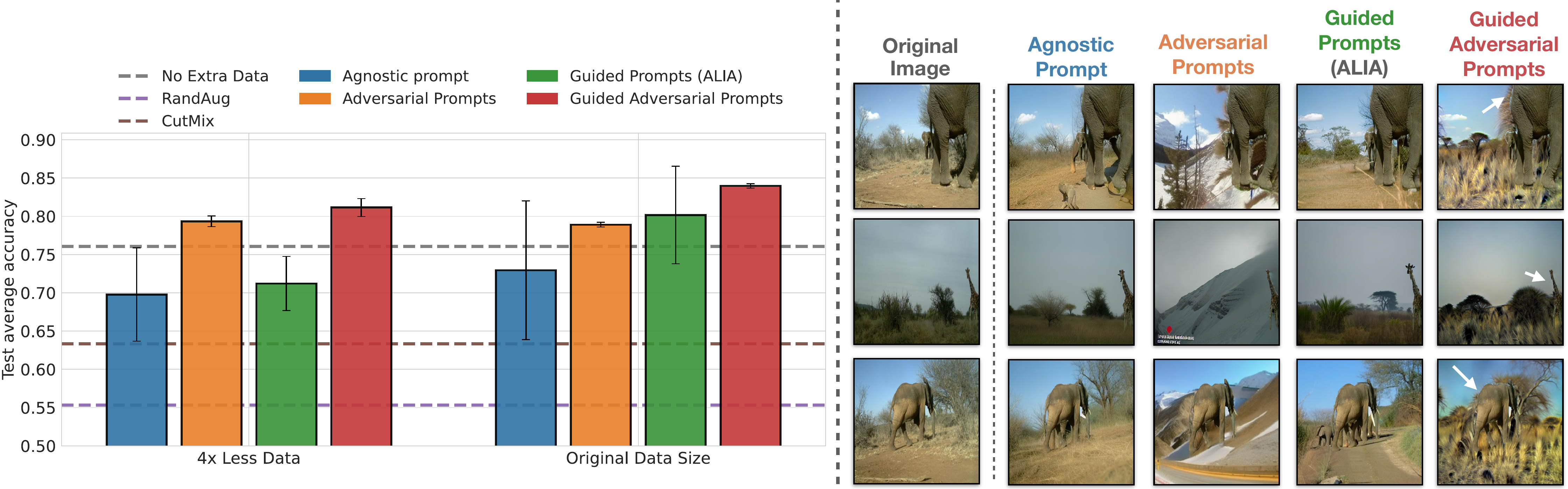} 
    % \vspace{-2mm}
    \caption{
    % \footnotesize
    \captiontitle{\gaplong{} generate hard training examples leading to the best performance.}
    % \textbf{Left:} \textbf{For iWildCam, \gaplong~are superior in performance and data-efficiency.} 
    % We train a model on the combination of the original training data and extra data generated using different types of prompts.  
    % We show the average accuracy on two iWildCam test camera trap locations.
    % We run each experiment with three seeds and report the mean and standard deviation.
    % \textbf{(1)} \seabornred{\gaplong}~consistently outperforms \seaborngreen{Guided Prompts} and \seabornorange{\aplong}.
    % For \seabornorange{\ap}~and \seabornred{\gap}, there is a much lower performance drop (1-3\%) when reducing the amount of generated data compared to \seaborngreen{GP}~(9\%). This suggests that being only target-informed requires a more exhaustive exploration of the image space to find samples that are "useful" for the pre-trained model, compared to being model-informed.
    \textbf{\textit{Left:}} Performance on the images from unseen camera locations (out-of-distribution) with different numbers of generated synthetic images for iWildCam.
    We repeat each experiment three times and report the mean and standard deviation.
    \seabornred{\gap}~consistently outperforms other types of prompts.
    We also find that \seabornorange{\ap} significantly outperforms \seaborngreen{\gp} in the low-data regime, suggesting that being only target-informed requires a more exhaustive sampling to find useful training examples.
    \textbf{\textit{Right:}} 
    Original training examples from iWildCam and generated examples for each type of prompt.
    For \seabornred{\gap}, arrows point to the animals for clarity; from top to bottom, they are misclassified as cattle, dik-dik, cattle. 
    \seabornred{\gap} generates images that are both visually coherent with the target location (described as \quotes{grassy field}) and introduce an adversarial \quotes{camouflage} effect.
    % See \cref{sec:classification} for more details.
    % \textbf{Qualitative results and comparison with ALIA on the iWildCam dataset.} From left to right: the real training data from iWildCam; generation using the \seabornblue{Agnostic Prompts} template: \quotes{\textit{a camera trap photo of \{class name\}}}; \seaborngreen{\gplong}~(ALIA) generations with target-informed prompt template: \quotes{\textit{a photo of \{class name\} in a grassy field with trees and bushes}}; \seabornorange{\aplong}~generations; and \seabornred{\gaplong}~generations with the same target distribution as \seaborngreen{\gplong}. \seabornred{\gap}~generations, while being visually coherent with the \quotes{grassy field} location, introduce an adversarial \quotes{camouflage} effect. White arrows point to the animals for clarity. From top to bottom, they are (mis)classified as cattle, dik-dik, cattle, cattle. 
    }
    \label{fig:iwild-joint}
    % \vspace{-5mm}
    % \vspace{-7mm}
\end{figure}

\textbf{iWildCam results.}
\cref{fig:iwild-joint} shows that similar to Waterbirds, \gap, combining both feedbacks, achieves the best performance.
Compared to GP with the target-only feedback, \gap{} generates images with a ``camouflage'' effect that fools the classification model.
Compared to AP with the model-only feedback that generates images with snow background, \gap's images are more similar to the target domain that is summarized as ``grassy field'' (by GPT-4 in~\citep{dunlap2023diversify}).
We also find that in the case of iWildCam, \ap~significantly outperforms \gp~in the low-data regime, suggesting that model-informed feedback generates more useful training examples.
We also find that~\ap \ and~\gap ~outperform standard (non-generative) augmentation baselines {and domain generalization baselines (see~\cref{app:iwild-dg} for results).}

% \cref{fig:iwild-joint} shows that having model-informed feedback (\aplong) helps to generate more useful data than no feedback mechanism (Agnostic Prompts) and also improves the performance of the target-only informed \gplong~method in the low-data regime.
% \gaplong~combines the benefits of both model- and target-informed feedback mechanisms, consistently outperforming other methods.
% \cref{fig:iwild-joint} shows exemplar generations for each method.
% We find that while \ap~generates images distinct from the target distribution and images generated by target-informed methods (snow background vs. grass background), training a model using these examples in the low-data regime performs better than \gp~and similar to \gap.

\subsection{Depth Estimation}
\label{sec:depth}
\begin{figure*}[t] 
\centering 
\vspace{-8mm}
\includegraphics[width=0.95\columnwidth]{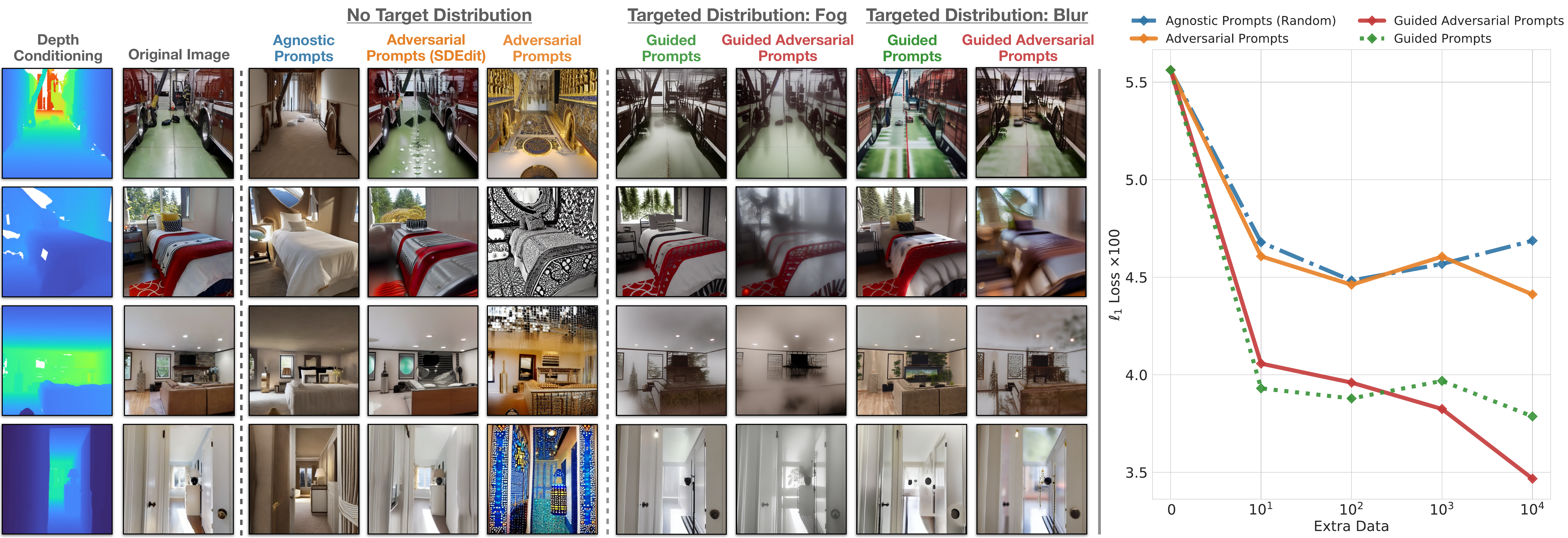}
\caption{
% \footnotesize
\textbf{Left: A comparison of generations with different prompts.} 
% A comparison of the generated data for the baselines and our method. 
Generations with \seabornorange{\ap}~results in \textit{diverse styles} that are \textit{distinct from the original training data and with agnostic prompts}. \seabornorange{\ap~(SDEdit)} generations have slight modifications to the original image, as it was conditioned on them.
% \seabornorange{\aplong}~with SDEdit (fourth column) are. As these generations are conditioned on the original image, they look more similar to them.
The last 4 columns show the generations with text guidance for target shifts \textit{fog} and \textit{blur}. 
% Using \seaborngreen{\gp}~alone, in this case ``fog'' or ``blur'', results in generations with a mild fog or blur. 
Compared to \seaborngreen{\gp}, \seabornred{\gap}~results in generations with more severe fog or blur.
\textbf{Right: Performance of \seabornred{\gap} with different amount of added data.} The target distribution here is \textit{defocus blur} applied to Taskonomy test images. The plot shows the $\ell_1$ loss of the UNet model versus the amount of extra data generated and used for fine-tuning. 
% In this example, we performed image guidance with a set of 100 \textit{unlabelled} corrupted RGB images. 
The plot shows that both \seaborngreen{\gp} and \seabornred{\gap} were able to guide the optimization toward generating training data relevant to the distribution shift. 
% Furthermore, there is a large improvement compared to using \seabornorange{\ap} or the baselines with as little as 10 extra generated samples. 
See~\cref{app-fig:depth-gen-comparison} for results on other corruptions {and~\cref{app:add-gen-data} for results on $10^5$ data}. % \ty{update visuals to include image guidance} 
}
\vspace{-5mm}
\label{fig:depth-gen-comparison}
\end{figure*}

\vspace{-1mm}
% We use Taskonomy~\citep{zamir2018taskonomy} as our training dataset which includes 4 million images of indoor scenes.
We use the Taskonomy~\citep{zamir2018taskonomy} and Omnidata~\cite{eftekhar2021omnidata} datasets for training. The former consists of 4 million images of indoor scenes, while the latter is a mixture of datasets and contains both indoor and outdoor scenes.
% \ty{mention taskonomy and omnidata dataset}
We evaluate our method on a range of domain shifts, from Common Corruptions~\citep{hendrycks2019benchmarking}, 3D Common corruptions~\citep{kar_3d_2022} to other datasets like Replica.
We compare the following methods. They all involve fine-tuning the originally trained model $f$ on different synthetically generated datasets. 
% We use ControlNet v1.0 with depth conditioning for the experiments in this section. 
See \cref{fig:depth-gen-comparison} for a comparison of the generations: 
\begin{itemize}[leftmargin=*, label={}] \vspace{-2mm}%[leftmargin=2.0mm,label={}]
    \setlength{\itemsep}{0.3pt}%
    \setlength{\parskip}{2pt}%
	\item \textit{Control (No extra data)}: We fine-tune $f$ on the original training data. This baseline is to ensure that the difference in performance is due to the generated data rather than, e.g., longer training or other hyperparameters used during fine-tuning.
    \item \textit{Agnostic Prompts}: This baseline generates data that is \textit{agnostic} to the model and the target distribution. We generate images with the prompt \quotes{\textit{room}} as the datasets consist of indoor images from mostly residential buildings.
    \item \textit{Agnostic Prompts (Random)}: 
    % We generate data with \quotes{random} prompts. 
    As we use 30 tokens for our methods, this baseline controls for the number of tokens used. Here, we sample the same number of tokens as \ap~randomly from the vocabulary to construct a prompt. See~\cref{app:depth-training-details} for details.
    % In our proposed method, 
    % we optimize for $n$ embedding vectors, resulting in a prompt, $c$. Thus, to match this setting, from a Gaussian distribution fitted on the embeddings from the vocabulary, we sample $n$ random embeddings to create a random prompt to be used in the data generation. \ty{shorten and rephrase}
    \item \textit{\aplong}: We perform the optimization as described in \cref{eq:adversarial-opt}, to get 30 adversarial prompts (tokens).
    % \todo{resulting in X adversarial prompts (tokens), generate data using these prompts} and fine-tune on it.
    \item \textit{\gplong}: We generate data using only the CLIP guidance loss described in \cref{eq:clip-guidance}. We consider both image and text guidnace. For the former we use about 100 \textit{unlabelled} images from the target distribution, for the latter we use the name of the shift e.g., ``fog''.
    \item \textit{\gaplong}: We combine adversarial and guidance losses. We use the same settings as \aplong~and apply CLIP guidance loss used for \gplong. This allows us to generate data that is both useful for the model and target distribution.
    % In addition to optimizing the adversarial loss, we also optimize the CLIP guidance loss as described in \cref{eq:adversarial-opt_and_clip-guidance}. This allows us to generate data that is also informed of a certain distribution shift.
\end{itemize}
\textbf{Training details.}\label{para:depth-train-dets}
We consider the following pre-trained models for $f$: {1)}~a U-Net~\citep{ronneberger2015u} model trained on the Taskonomy dataset~\citep{zamir_taskonomy_2018,zamir_robust_2020} and  {2)}~a dense prediction transformer~(DPT)~\citep{ranftl_vision_2021} model trained on Omnidata~\citep{eftekhar_omnidata_2021}. The adversarial optimization was done with AdamW~\citep{loshchilov_decoupled_2019}, learning rate of $5.0\times 10^{-4}$, weight decay of $1.0 \times 10^{-3}$. We set the early stopping threshold (mentioned in \cref{sec:alignment-collapse}) to 0.08 for the UNet model and 1.0 for the DPT model. {They were trained with $\ell_1$ and Midas loss~\citep{eftekhar_omnidata_2021} respectively.} 
We perform a total of 30 runs to get 30 different \aplong. 
Similar to classification, for efficiency, we use 5 denoising steps during optimization and 15 for the final generations, with the DDIM~\citep{song2020denoising} scheduler, {which we found to be a good trade-off between quality and efficiency.}
See the~\cref{app:depth-training-details} for more details.
% During optimization, we use only 5 denoising steps, as it is more stable.
% For Guided Adversarial Optimization, the guidance coefficient for text and image guidance is 1 and 5 respectively. 
% For fine-tuning, we generate images with 15 steps. For the GP runs with SDEdit, we used strength 0.6, for the GAP runs, strength 0.9. See the~\cref{app:depth-training-details} for further details.

\textbf{Comparing the generated images with different prompts.} 

\begin{wraptable}{r}{0.6\textwidth} %{0.55\linewidth} 
    \vspace{-4mm}
    \caption{
    % \footnotesize 
    \captiontitle{Quantitative results for depth estimation.}
    $\ell_1$ errors for the depth prediction task for U-Net and DPT models on Common Corruptions (CC), 3D Common Corruptions (3DCC), and cross-datasets~(CDS) like Replica. 
    Results for CC and 3DCC are averaged over all distortions and severity levels. 
    % (Lower is better. U-Net losses are multiplied by 100 and DPT losses by 10, for readability).
    % Note that performance between models is not comparable as they were trained to output absolute and relative depth, respectively.
    % (see~\cref{para:depth-train-dets}), 
    % $\ell_1$ for the former and Midas loss~\citep{eftekhar2021omnidata} for the latter, 
    % We evaluate on distribution shifts from Common Corruptions (CC), 3D Common Corruptions (3DCC), and cross-datasets~(CDS) like Replica.
    \ap~outperform baselines for all models and distribution shifts.
    % Our method is able to generate training data that can improve results over the baselines on several distribution shifts. 
    % Performing \seabornorange{\ap} with SDEdit gives better results than \seabornorange{\ap} under distribution shifts. 
    % Thus, conditioning on the original image seems to be helpful for these shifts. 
    % For the DPT model, the trends are similar, \seabornorange{\ap} performs better than the baselines. 
    As \gap~requires optimizing for each shift individually, and the benchmarks contain $\sim30$ shifts in total, we only perform {\gap} for a few shifts (see~\cref{fig:depth-gen-comparison} and~\cref{app-fig:depth-gen-comparison} for results). See~\cref{app-tab:depth-results} for results on other baselines.}
    \begin{adjustbox}{width=0.58\textwidth}
    \begin{tabular}{c|cccc|cc} \toprule
                                 & \multicolumn{4}{c|}{UNet $(\times100\downarrow)$}               & \multicolumn{2}{c}{DPT $(\times10\downarrow)$}                 \\ \cmidrule{2-7}
                                 & \multicolumn{3}{c}{Taskonomy} & Replica & \multicolumn{2}{c}{Taskonomy} \\ \midrule
    Shift                        & Clean     & CC      & 3DCC    & CDS     & CC            & 3DCC              \\ \midrule
    Control (No extra data)                     & \textbf{2.35}      & 4.93    & 4.79    & 5.38    & 3.76          & 3.42              \\ \midrule
    \seabornblue{Agnostic Prompts}             & 2.47      & 5.03    & 4.17    & 5.30    & 4.06          & 3.58              \\
    \seabornblue{Agnostic Prompts} (Random)    & 2.38      & 4.96    & 4.11    & 5.14    & 3.88          & 3.51              \\
    \seabornorange{\aplong}          & 2.49      & 4.36    & 4.02    & 5.12    & 3.40          & 3.28              \\
    \seabornorange{\aplong} (SDEdit) & 2.59      & \textbf{4.20}    & \textbf{3.88}    & \textbf{4.96}    & \textbf{3.35}             & \textbf{3.25}                    \\ \bottomrule
    \end{tabular}
    \end{adjustbox} \label{tab:depth-results} \vspace{-2mm}
  \end{wraptable}

\cref{fig:depth-gen-comparison}-left shows the generations from the baselines and our methods. 
% The generations with Agnostic Prompts tend to have similar styles. 
The generations with \ap~are diverse and have complex styles. Incorporating SDEdit results in slight modifications to the original image, as it was used for conditioning. 
% As expected, using SDEdit during the optimization results in generations closer to the original image.
The last four columns show the results of using text guidance for the target distribution shift \textit{fog} and \textit{blur} (see~\cref{app:clip-img-vs-text} for a comparison between generations from image and text guidance). 
% \todo{reduce caption or reduce this paragraph}
% , as described in \cref{sec:clip-guidance}. 
% , with and without adversarial optimization.
% The generations with Guided Prompts involve passing the depth conditioning and prompt ``fog'' or ``blur'' to the diffusion model. 
The generations for \gp~results only in a mild level of fog or blur.
In contrast, \gap~results in more severe fog and blur corruptions. {The generations for \gp, using only the guidance feedback, result only in a mild level of fog or blur.
Finally, incorporating both feedback mechanisms in \gap~results in generating images with corruptions that are both relevant to the corresponding target distribution and more severe, making them more useful for improving the model.
}
% See~\cref{app:clip-img-vs-text} for generations using image guidance.
% These generations were attained with text guidance. 

% In~\cref{app:clip-img-vs-text}, we compare the qualitative and quantitative differences from image and text guidance. Text guided prompts tends to generate corruptions that are more realistic (see~\cref{app-fig:depth-gen-comparison} for comparisons). However, image guidance tends to perform better quantitatively across the different target distributions. 
% We provide generations using CLIP image guidance in~\supmat~\cref{app:clip-img-vs-text}.
% Note that all the generations follow the conditioning, i.e., depth labels (see first column).
% See~\cref{sec:adv-prompt} for the discussion on how we prevent the generations from collapsing.
% Thus, this gives us \textit{aligned training data}, i.e., RGB images and depth labels that we can use for fine-tuning. 

\textbf{Quantitative results.} 
Tab.~\ref{tab:depth-results} shows the results from finetuning our method and the baseline on the different generated datasets. \aplong~improves the performance of the model under different distribution shifts for both the UNet and DPT models.
% We evaluate our method and the baselines after fine-tuning on their respective generated datasets. Tab.~\ref{tab:depth-results} demonstrated that data generated with \aplong~improve the performance of the model under different distribution shifts.
% Furthermore, we show that the trend holds with the DPT model.
% Thus, our method is able to successfully find useful \aplong~for different architectures. 
% \textbf{Performance of \gap~against amount of generated data.} 
\cref{fig:depth-gen-comparison}~\figright~shows the results with \gaplong~for the \textit{defocus blur} corruption against the amount of extra generated data used for finetuning.
\gplong~and \gaplong~result in a large performance improvement, compared to \aplong~or the baseline with only 10 extra data points. 
This suggests that the guidance loss \textit{successfully steered the generations toward producing training data relevant to the distribution shift}. See~\cref{app-fig:depth-gen-comparison} for results on other corruptions.
{As the dataset size increases, \gp~shows signs of saturation, whereas \gap~further improves the model, showing the benefit of the adversarial feedback mechanism.}
% \ty{move some sentences from caption here}
% This experiment was performed with image guidance. 

% Unlike classification, the performance of \gap~for depth is not consistent across all distribution shifts. 
% % The diffusion model was not able to generate certain shifts e.g., noise corruptions, as the text descriptions and unlabelled images was too ambiguous e.g., `noise' or having common attributes other than the corruption. 
% The diffusion model was not able to generate certain shifts e.g., noise corruptions, as the text descriptions e.g., `noise', was too ambiguous.
% We leave further analysis to future work.

% \clearpage
\subsection{Additional Analysis}
\label{sec:analysis}

\begin{wraptable}{r}{0.5\textwidth} 
\vspace{-4mm}
\centering
\caption{
% \footnotesize 
\textbf{Transferability of adversarial prompts.}
We show the results of different combinations of the finetuning model and model used to generate additional data via model-informed feedback.
We report top-1 accuracy for iWildCam and $l_1$ loss for depth.
% \ty{i think if we move this table to right beside the text we dont need to explain too much?}
} \vspace{-2mm}
\label{tab:analysis-feedback}
\resizebox{0.48\textwidth}{!}{%
% \begin{tabular}{l|ll||l|ll} \toprule
% % \multicolumn{3}{c||}{iWildCam} & \multicolumn{3}{c}{Depth} \\ \midrule
% \textbf{iWildCam}  & \multicolumn{2}{c||}{Finetuned model~($\uparrow$)} &  \textbf{Depth} & \multicolumn{2}{c}{Finetuned Model~($\downarrow$)} \\ \midrule
% Data from & \multicolumn{1}{c}{ResNet} & \multicolumn{1}{c||}{ViT} & Data from & UNet & DPT \\ \midrule
% Agnostic Prompts & 72.94 & 73.07 & Agnostic Prompts & 5.27 & 3.66 \\ %\midrule
% ResNet & 83.97 & 72.61 & UNet & 4.73 & 3.39 \\
% ViT & 83.87 & 77.21 & DPT & 4.88 & 3.46 \\ \bottomrule
% \end{tabular}
\begin{tabular}{@{}lcc||lcc@{}}
\multicolumn{3}{c}{iWildCam, Acc. ($\uparrow$)}                                        & \multicolumn{3}{c}{Depth, $l_2$ $(\downarrow)$}                                        \\
\toprule
                                                           % & \multicolumn{2}{c||}{Model} & \multicolumn{1}{c}{}                                       & \multicolumn{2}{c}{Model} \\
Data\textbackslash Model                                                   & ResNet       & ViT         & Data\textbackslash Model                                                   & UNet        & DPT         \\ \midrule
ResNet                                                     & \textbf{83.97}        & 72.61       & UNet                                                       & \textbf{4.73}        & \textbf{3.39}        \\
ViT                                                        & 83.87        & \textbf{77.21}       & DPT                                                        & 4.88        & 3.46        \\ \midrule
\begin{tabular}[c]{@{}l@{}}Agnostic\\ Prompts\end{tabular} & 72.94        & 73.07       & \begin{tabular}[c]{@{}l@{}}Agnostic\\ Prompts\end{tabular} & 5.27        & 3.66        \\ \bottomrule
\end{tabular}
}
\label{tab:model_generalization_iwild}
\vspace{-2mm}
\end{wraptable}

% \textbf{Fine-tuning with \gaplong~from a different model.}
% \textbf{Are \aplong~from another model useful for fine-tuning?}
% \textbf{Do different models need different data?}
% \textbf{Are adversarial prompts found for one model useful for training another model?}
\textbf{Are adversarial prompts found for one model specialized for that model?}
We study 1) whether models benefit \textit{the most} from data generated using their own feedback and 2) whether they benefit at all from using data generated for another model (compared to Agnostic Prompts.) 
% We explore whether models benefit more from data generated from using their own feedback and whether they benefit from using data generated for another model compared to Agnostic Prompts. 
% We study this for iWildCam and depth. For the former, we finetune a ResNet model on \gap~data obtained from ViT and vice versa. For the latter, we finetune a UNet model on \ap~data obtained from DPT and vice versa. 
% We also fine-tuning on agnostic prompts for comparison.
\cref{tab:analysis-feedback} shows that in most cases, a model performs best when trained with data generated using its own feedback, which shows the usefulness of the model-informed feedback mechanism.
% \todo{In \cref{tab:app-analysis-feedback-depth}, we show that the DPT case is not due to the failure of adversarial optimization, which does find adversarial prompts for that particular model.}
At the same time, we find that in most cases, prompts found for another model still outperform Agnostic Prompts, suggesting that adversarial prompts result in useful transfer between models.

\begin{wrapfigure}{r}{0.5\textwidth} \vspace{-9mm}
% \begin{figure}
    \includegraphics[width=0.49\columnwidth]{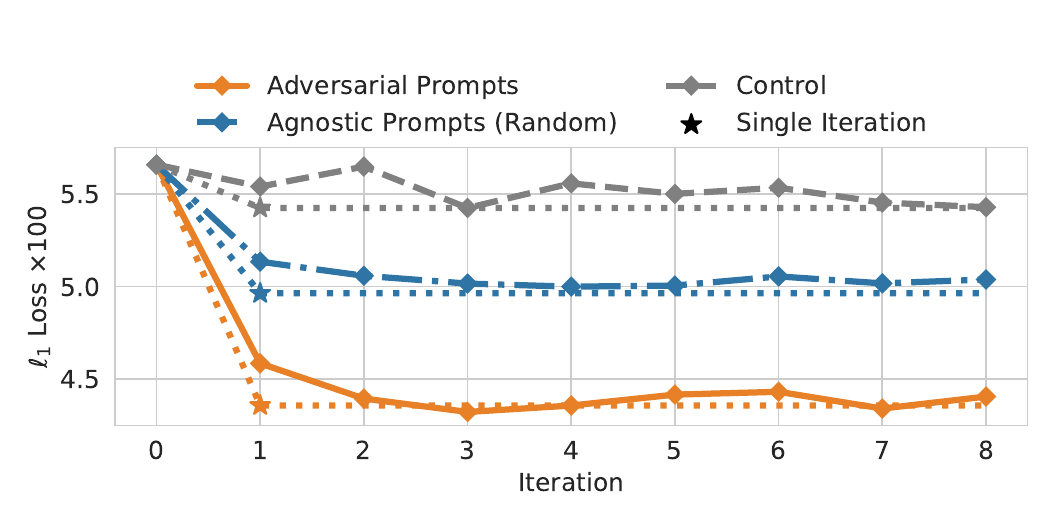} \vspace{-3mm}
    \caption{
    % \footnotesize
    \textbf{Comparing the performance from running multiple iterations versus a single iteration} of adversarial optimization, generation, and fine-tuning. The plot shows the $\ell_1$ loss ($\times 100$) of the U-Net model against the number of iterations on the depth prediction task. The loss is computed on the Taskonomy dataset under common corruptions, averaged over all corruptions and severity levels. 
    % See~\cref{app:multi-iter} for further details. 
    % The first iteration of the multi-iteration run resulted in the largest improvement in performance. However, it converges to a similar performance as the single iteration run. Thus, we chose to perform only a single iteration in \cref{tab:depth-results}.
    }  
    \label{fig:depth-multi-iter} \vspace{-3mm}
% \end{figure}
\end{wrapfigure}
% \textbf{Running multiple iterations of adversarial optimization vs a single iteration.}
\textbf{Does multiple iterations of adversarial optimization further improve performance?}
We define an \textit{iteration} as one round of adversarial optimization, i.e. optimizing \cref{eq:adversarial-opt} or \cref{eq:adversarial-opt_and_clip-guidance}, generation and fine-tuning.
Given that all of the above results were obtained with a single iteration, we aim to see if there are benefits in performing multiple iterations. 
% We perform a total of 8 iterations and we compare this to performing a single iteration.
% The experimental settings for 8 iterations and a single iteration are similar, e.g., in total, over the 8 iterations, we optimize for the same number of \aplong, and fine-tune on the same datapoints, etc.
We perform a total of 8 iterations keeping the total number of generated images, prompts, etc., the same as for a single iteration for comparison.
% The single- and multi-iteration runs have similar settings in terms of the total number of \seabornorange{\ap} used, the same number of tokens per prompt, etc. 
\cref{fig:depth-multi-iter} shows that the first iteration of the multi-iterations setting results in the largest performance improvement and eventually converges to the performance of the single iteration approach.
Thus, we chose to perform a single iteration for our main experiments.
See the~\cref{app:multi-iter} for additional implementation details and results. 
% \ty{TODO: mention curriculum learning? maybe mention that we expect this single iteration being enough phenomenon to hold with "strong/competent" models, but if we start a weak model, it's probably beneficial to do multiple rounds, thus implicitly doing curriculum learning}

% \ty{TODO: add para on guidance/loss ablation from app?}
\section{Conclusion and Limitations}
\label{seq:conclusion-limit}
In this work, we aim to generate training data useful for training a supervised model by steering a text-to-image generative model.
We introduced two feedback mechanisms to find prompts informed by both the given model and the desirable target image distribution.
Evaluations on a diverse set of tasks and distribution shifts show the effectiveness of the proposed closed-loop approach in comparison to open-loop ones.
Below we discuss some of the limitations of our work:
\begin{itemize}[leftmargin=*,label={}] \vspace{-1mm}
    \setlength\itemsep{0.3pt}
    \item \textit{Label shift:} 
    In this work, we focus on generating novel images.
    However, some distribution shifts can also change the label distribution, e.g., for depth estimation, changing from indoor to outdoor scenes would result in a shift in depth maps.
    One possible approach could be learning a generative model over the label space~\citep{le2021semantic} to control the generation in both the label and image space.
    \item \textit{Computational cost:} 
    Estimating the gradient of the loss in \cref{eq:adversarial-opt_and_clip-guidance} requires backpropagation through the denoising process of the diffusion model, which can be computationally demanding.
    Using approaches that reduce the number of denoising steps~\citep{song2020denoising,luo_latent_2023} may be able to reduce this computational cost.
    \item \textit{Label Conditioning:} 
    As discussed in \cref{sec:adv-prompt}, our method is limited by the faithfulness of the generation conditioned on the given label.
    For example, we found that the semantic segmentation ControlNet does not follow the conditioning accurately enough to be useful for the supervised model.
    Further developments in more robust conditioning mechanisms are needed to successfully apply our method to other tasks.
\end{itemize}

\bibliography{main}
\bibliographystyle{tmlr}

\appendix
% \maketitlesupplementary

% \twocolumn[
%     \centering
%     \Large
%     \textbf{\thetitle}\\
%     \vspace{0.5em}Appendix \\
%     \vspace{1.0em}
% ] %< twocolumn

\section{Appendix Outline}
We provide further discussions, details, and evaluations in the appendix, as outlined below.

\begin{itemize}
    \item \cref{app:classification,app:depth} describe additional implementation details for our classification and depth estimation experiments, respectively.
    \item \cref{sec:ti-guidance} describes an image guidance mechanism using Textual Inversion~\citep{gal2022image} and compares it with the CLIP guidance mechanism, on Waterbirds.
    \item 
    % \cref{app:depth-augs} provide additional results for ``standard'' augmentation baselines for the depth estimation experiments.
    {\cref{app:depth-iwild-baselines} provides additional quantitative results. See \cref{app:depth-augs} for comparisons with other data augmentation baselines for depth estimation and \cref{app:iwild-dg} for comparisons with domain generalization baselines on iWildCam. See \cref{app:add-gen-data} for results on training with more synthetic data.}

    \item \cref{app:waterbirds-additional-qual,app:iwild-additional-qual,app:depth-additional-qual} provide \textbf{qualitative generations from all the \aplong~and \gaplong}~used in the Waterbirds, iWildCam, and depth estimation experiments. 
    \begin{itemize}
        \item     Additionally, for depth estimation, we provide a qualitative comparison of \aplong~generations optimized on different models (UNet~\citep{ronneberger2015u}, DPT~\citep{ranftl_vision_2021}). For iWildCam, we also provide additional results using a ViT-B-16~\citep{dosovitskiy2020image} instead of a ResNet50~\citep{he_deep_2016}.
        % \hbtodo{add ViT}
    \end{itemize}

    \item \cref{app:depth-additional-analysis} provides additional analysis on the depth estimation experiments: 
    \begin{itemize}
        \item the single iteration vs. multi-iteration setting
        \item a comparison of CLIP image and text guidance
        \item \textbf{an assessment of the generalization of \aplong~from one model to another.}
    \end{itemize}

\end{itemize}

\section{Implementation Details for Classification Tasks}
\label{app:classification}

\subsection{Training data generation}

\paragraph{Inpainting.}
\label{sec:inpainting} %Sec.~3.1
As mentioned in main paper \cref{sec:prelim}, for semantic classification tasks, we utilize the foreground object masks and use an in-painting technique proposed in \cite{lugmayr_repaint_2022} that preserves the masked region throughout the denoising process. 
In this section, we briefly describe this procedure and refer the reader to the original work for more details.
% We describe here in more details how we use inpainting. 

Let $m$ be a binary pixel mask, where a pixel is equal to 1 if the pixel contains the object and 0 otherwise, and $x$ be the original image from a training dataset.
During generation, after obtaining a denoised sample $\xsample_{t}$ at time $t$ we update it as $\xsample_t \leftarrow m \odot x^\t{orig}_{t} + (1-m)\odot \xsample_{t}$, where $x^\t{orig}_{t}$ is the original image noised to have the correct properties of the expected Gaussian distribution at time $t$.
%and $x_{t-1}$ is the result of the usual denoising step from $x_t$ to $x_{t-1}$.

However, because we are using \SD~\cite{rombach2022high}, the denoising process is done in latent space (using an encoder $\mathcal{E}$), not pixel space. This means that to apply inpainting, we must resize the mask $m$ to the latent space dimensions, and apply the above-described procedure in the latent space: $\Tilde{z}_t \leftarrow m_z \odot z^\t{orig}_{t} + (1-m_z)\odot \Tilde{z}_{t}$, where $z^\t{orig}_0 = \mathcal{E}(x^\t{orig})$ and $z^\t{orig}_t$ is its corresponding noised version.
While this procedure usually performs well in preserving the original region of interest, we also paste the original masked region in the pixel space to obtain the final sample $\xsample = m \odot x^\t{orig} + (1-m)\odot \xsample_0$.
% Given this is an approximation, it sometimes does not preserve adequately the object mask.
% Thus, in addition to inpainting, given $\xsample_0$, the final output of the denoising process, and the original image $x^\t{orig}$ we also paste the original object mask on the final output, and our final sample is $\xsample = m \odot x^\t{orig} + (1-m)\odot \xsample_0$.

\paragraph{SDEdit~\cite{meng2021sdedit}.} In addition to inpainting, depending on the setting, we also use SDEdit~\cite{meng2021sdedit}, a mechanism available to all diffusion models that allows to use an initial image to condition the generation of new images to be closer to the initial image. The mechanism is parametrized by the \textit{SDEdit strength} $s$, which indicates the extent by which the model can deviate from the original image. 

\paragraph{Text-to-image model.}
For our diffusion model, we use Stable Diffusion v1.5~\footnote{https://huggingface.co/runwayml/stable-diffusion-v1-5}.

\subsection{ALIA}
\label{sec:alia}
Here, we give more details on the ALIA~\cite{dunlap2023diversify} (Automated Language-guided Image Augmentation) baseline method, which aims at generating images targeting a particular test distribution similar to our guidance mechanism (main paper \cref{sec:clip-guidance}). %Sec.~3.3

Given exemplar images from the test distribution, ALIA first captions each image using the BLIP~\cite{li_blip_2022} captioning model.
% This produces a comprehensive set of captions.
Then, it uses the GPT-4~\cite{openai_gpt-4_2023} LLM to summarize these captions into a list of domains asking it to produce descriptions that are agnostic to the class information.
\cite{dunlap2023diversify} then use these prompts to generate additional training data.
In order to preserve the original class information in their generations, they use SDEdit~\cite{meng2021sdedit} or Instruct Pix2Pix~\cite{brooks2023instructpix2pix}.
We refer the original paper for further implementation details.
Below, we summarize resulting prompts we use for comparison in our results.

For Waterbirds~\cite{sagawa2019distributionally}, we found that removing the prefix \textit{“a photo of a \{class name\}"} from the original prompts when using the inpainting technique (\cref{sec:inpainting}) to work slightly better for both the ALIA baseline and our CLIP text guidance (main paper \cref{eq:clip-guidance}).
We, therefore, use the following prompts:
\begin{itemize}
    \item \textit{“in a bamboo forest with a green background.”}
    \item \textit{“flying over the water with a city skyline in the background.” }
    \item \textit{“perched on a car window.”}
    \item \textit{“standing in the snow in a forest.”, }
    \item \textit{“standing on a tree stump in the woods.”}
    \item \textit{“swimming in a lake with mountains in the background.”,}
    \item \textit{“standing on the beach looking up.” }
\end{itemize}

For iWildCam~\cite{beery2021iwildcam}, we keep the original prompts intact: 
\begin{itemize}
    \item \textit{“a camera trap photo of a \{class name\} in a grassy field with trees and bushes.” }
    \item \textit{“a camera trap photo of a \{class name\} in a forest in the dark.”}
    \item \textit{“a camera trap photo of a \{class name\} near a large body of water in the middle of a field.” }
    \item \textit{“a camera trap photo of a \{class name\} walking on a dirt trail with twigs and branches.”}
\end{itemize}

\vspace{4mm}
There are two main differences between ALIA and our method:
\begin{enumerate}
    \item \textbf{The target distribution feedback}. ALIA aligns its prompts with the target distribution by utilizing captioning and summarizing. However, this summarizing process is not informed of the produced generations when using such prompts, and, thus, does not guarantee that the text prompt will accurately guide the generation process to images related to the target distribution.

    \item \textbf{Model feedback.} ALIA is not model-informed. Thus, it doesn't necessarily generate images \textit{useful} for training a given model. 
% However, it lacks the mechanism to refine the prompt based on the generated images to account for a potential semantic misalignment between the text prompt found by the language model and the text guidance of the generative model.
\end{enumerate}

\vspace{2mm}

Those two differences originate from the fact that ALIA is an \textbf{open-loop} method, i.e, it lacks the mechanism to refine the prompt based on the generated images. In contrast, our method uses model and target distribution feedback in a \textbf{closed-loop}. This allows our method to outperform ALIA and be more data-efficient.

\subsection{General implementation details}

\begin{table}[t]
\centering
\caption{Generation and optimization parameters for \aplong and \gaplong for classification tasks in \cref{sec:classification}.}
\resizebox{0.99\columnwidth}{!}{%
\begin{tabular}{@{}c|cccccc@{}}
\toprule
Dataset    & \multicolumn{1}{c|}{SDEdit strength} & \multicolumn{1}{c|}{denoising steps} & \multicolumn{1}{c|}{guidance scale} & \multicolumn{1}{c|}{\# placeholder tokens} & \multicolumn{1}{c|}{Opt. steps (AP/GAP)} & \multicolumn{1}{c|}{$\lambda_t / \lambda_i $} \\ \midrule
Waterbirds & 1.                                   & 5                                   & 7.0                                 & 5                                        & 1000/1000                                & 20/0                                          \\
iWildCam   & 0.8                                  & 5                                   & 5.0                                 & 10                                       & 2000/10000                               & 0/10                                          \\
\bottomrule
\end{tabular}
}
\label{tab:adv-opt-parameters} 
% \vspace{-5mm}
\end{table}

\begin{table}[t]
\centering
\resizebox{0.97\columnwidth}{!}{%
\begin{tabular}{@{}c|ccc|ccc@{}}
\toprule
Dataset    & \multicolumn{3}{c|}{ALIA} & \multicolumn{3}{c}{Ours} \\
 & \textit{SDEdit strength} & \textit{sampling steps} & \textit{text guidance} & \textit{SDEdit strength} & \textit{sampling steps} & \textit{text guidance} \\ \midrule
Waterbirds &  0.3     & 50       & 7.0     &  1.       &  15     &    7.0    \\    
iWildCam   & 0.5     & 50     & 7.5    & 0.8     & 5     & 5.0    \\ \bottomrule
\end{tabular}%
}
\caption{Generation parameters for the classification experiments}
\label{tab:classification-generation-parameters}
\end{table}

\begin{table}[t]
\centering
\begin{tabular}{@{}c|ccc@{}}
\toprule
Dataset    & \multicolumn{3}{c}{Training}                                     \\
           & \textit{learning rate} & \textit{weight decay} & \textit{epochs} \\ \midrule
Waterbirds & 0.001                  & 1e-4                  & 100             \\
iWildCam   & 0.0001                 & 1e-4                  & 100/20             \\ \bottomrule
\end{tabular}%

\caption{Training parameters for the classification experiments}
\label{tab:classification-training-parameters}
\end{table}

\paragraph{Generation.} We report our generation parameters in \cref{tab:classification-generation-parameters}. We use the DDIM~\cite{song2020denoising} scheduler. We generate 384x384 resolution images. Those parameters were chosen based on visual inspection, ease of optimization and downstream performance (validation accuracy).
\label{app:alia_filtering}
\textbf{Training data.}
After generation, ALIA's method consists of an additional filtering step to remove \quotes{bad} generations. This step relies on using a pretrained model to measure confidence on the generated images. However, given our method creates images that are adversarial to an iWildCam pretrained model, the filtering part of ALIA's pipeline is not usable on our data. Thus, to keep things comparable, we decided not to apply filtering both our method generated data and ALIA's generated data. However, it must be noted that \cite{dunlap2023diversify} only reports a 2\% absolute accuracy drop between fitlering and no filtering on iWildCam (1.4\% on Waterbirds), thus we do not expect a big difference in performance with ALIA's reported results and our results.

\paragraph{Supervised Training.} We report our training parameters in \cref{tab:classification-training-parameters}. We use ALIA's codebase to finetune our models, which ensures fair comparison to the ALIA baselines. For everything except the generated data, the settings are the same as in ALIA. For both datasets, the starting model is a ResNet50~\cite{he2016deep} model, pretrained on ImageNet~\cite{deng2009imagenet}.

The reported test accuracy is chosen according to the best checkpoint, measured by validation accuracy. 

%\todo{add caveat that ALIA previous code had a small bug where the test accuracy could update the value that tracked the best accuracy ?}. 

\subsection{Waterbirds}

\subsubsection{Dataset details}
\begin{figure}[t]
    \centering
    \includegraphics[width=0.48\columnwidth]{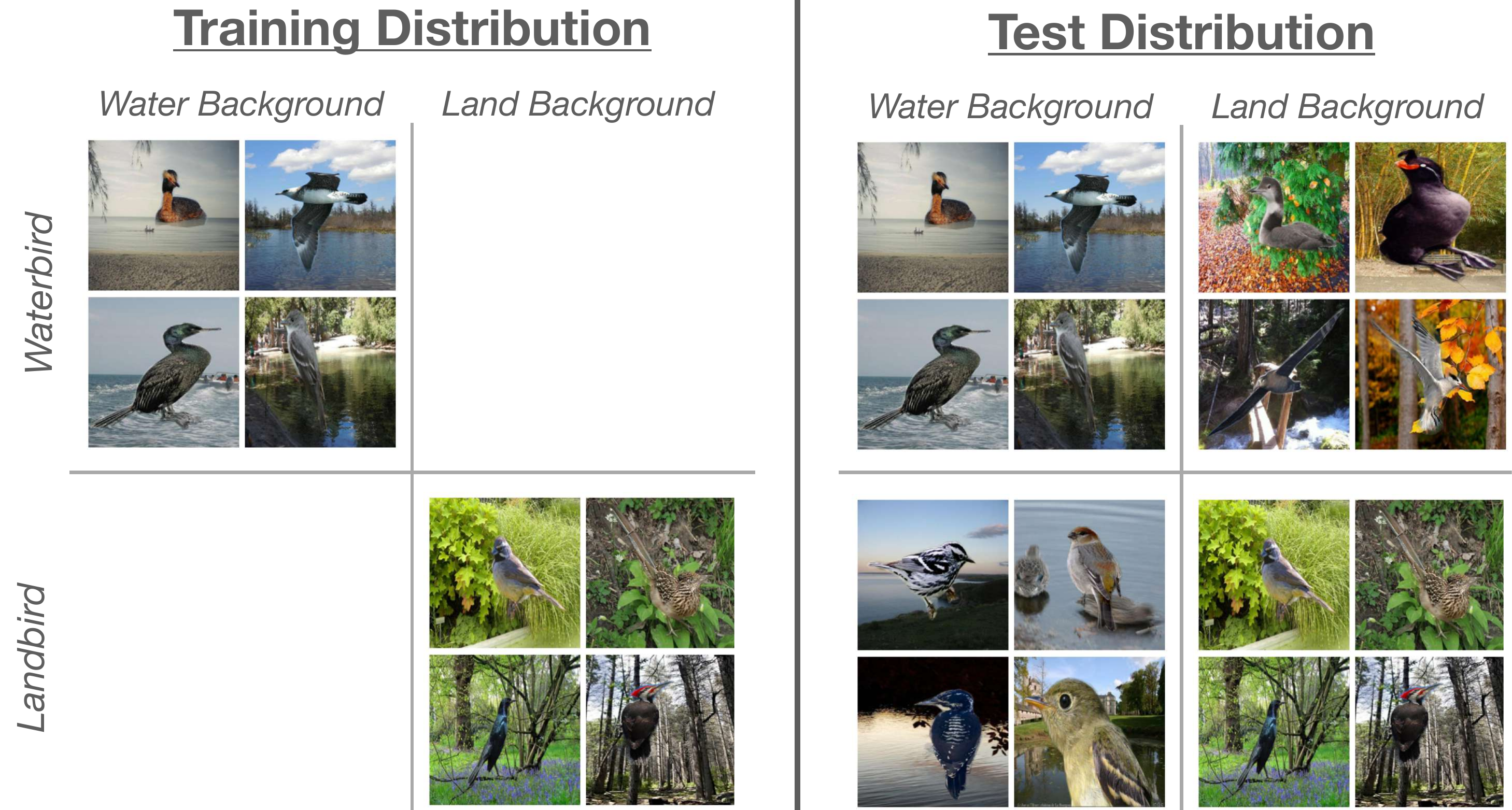}
    \caption{
    \captiontitle{Distribution shift in the Waterbirds dataset.} The background is a perfectly predictive spurious feature on the training distribution, but loses predictive power in the test distribution.
    }
    \label{fig:waterbirds-dataset}
\end{figure}

\cref{fig:waterbirds-dataset} demonstrates the shift between train and test distributions in the Waterbirds dataset \cite{sagawa2019distributionally}.
We follow the setting suggested in \cite{dunlap2023diversify} and use 1139 images as $\D_\t{tr}$, where waterbirds appear only on water background and landbirds on land background.
We add additional 839 examples either from the original dataset, where waterbirds appear only on land background and landbirds on water background ("Real OOD data"), or generated by \SD~with prompts obtained by one of the methods.
For the data-efficiency plots (e.g., \cref{fig:wb-results}) we reduced the number of added examples by a factor of \texttt{\{1/2, 1/4, 1/8, 1/16\}}.

Since the original Waterbirds dataset does not provide masks for the exact generated images, we used the SAM~\cite{kirillov_segment_2023} segmentation model to obtain bird segmentation masks for training images.
We use these masks to condition the generative model on the class by using inpainting as described in \cref{sec:inpainting}.

\subsubsection{Implementation Details}
\label{app:waterbirds-implementation-dets}

\paragraph{Adversarial Optimization.}
For adversarial feedback, we use the model trained only using the original training data $\D_\t{train}$ with complete spurious correlation. It is taken from ALIA checkpoints\footnote{https://api.wandb.ai/files/clipinvariance/ALIA-Waterbirds/y6zc932x/checkpoint/ckpt-Waterbirds-none-filtered-resnet50-1-0.001-0.0001/best.pth}.
As the task is the binary classification, we use the cross-entropy loss for the opposite class as the adversarial loss: $\L_\t{adv}(\xsample, y) = \L_\t{x-ent}(f(\xsample), 1 - y)$, assuming $y \in \{0, 1\}$. This is equivalent to the negative cross-entropy loss referred in the text.
We find four prompts per each class, i.e., eight prompts in total.
Each prompt is composed of five new learnable tokens.
We perform adversarial optimization for 1000 steps with learning rate 1e-3 using Adam~\cite{kingma_adam_2017}. 
We use five denoising steps during adversarial optimization and generate images for training with 15 steps.
We do not use SDEdit for Waterbirds.
See \cref{tab:classification-generation-parameters} for summary.
% \todo{report number of steps used, if different from generation}

\paragraph{CLIP Guidance.}
For Waterbirds, we use CLIP text guidance by encoding each of ALIA's summarized prompts (see \cref{sec:alia}) with the CLIP text encoder as described in main paper \cref{sec:clip-guidance}. %Sec.~3.3.
In addition, we renormalize the averaged target text embedding to have the norm equal to the mean norm of the original prompts, and use the resulting vector as the target $e_\t{t}$.
We use $l_2$ guidance loss: $\L_\t{t}(E_\t{t}(c_w), e_\t{t}) = \|E_\t{t}(c_w) - e_\t{t}\|_2^2$.
We use $\lambda_\t{t} = 20$ and $\lambda_\t{i} = 0$ (i.e., no image guidance).

% \subsubsection{Standard Augmentations Baseline}
% \label{app:waterbirds-augs}

% In \cref{fig:waterbirds-standard-augs}, we provide results with additional ``standard'' augmentation baselines.

\subsubsection{Additional Qualitative Results.}
\label{app:waterbirds-additional-qual}

\begin{figure*}[t]
    \includegraphics[width=\textwidth]{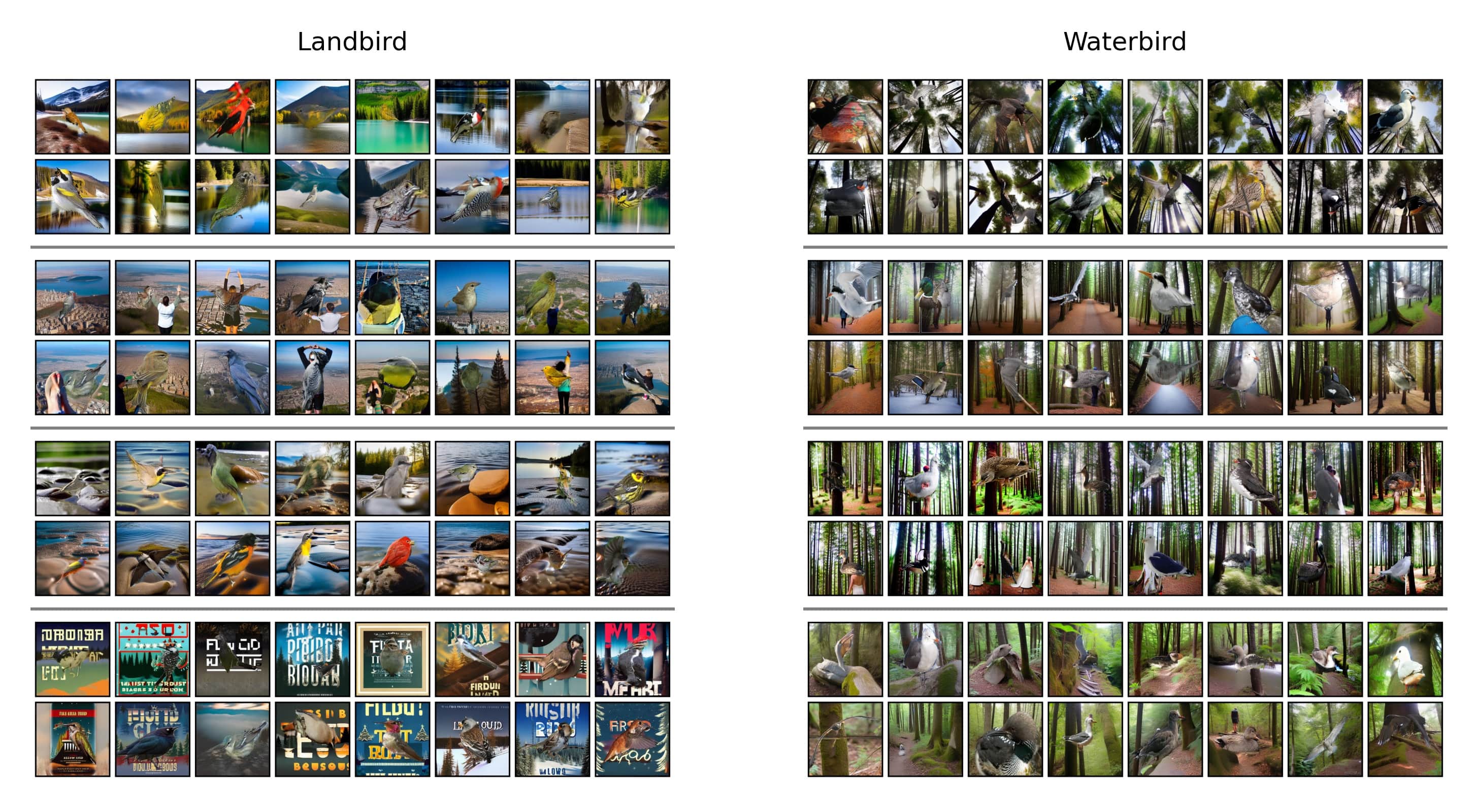}
    \caption{
    \captiontitle{Generation examples for \seabornred{\gaplong}}.
    Each column shows generations for the corresponding class.
    Each row in a column (separated by gray lines) shows generations for one found token.
    There are eight tokens in total, four for each class.
    % The generated images are samples that the pretained model fails on, while being close to the target distribution.
    \seabornred{\gap} tend to generate landbirds on water (\textbf{left}) and of waterbirds on land (\textbf{right}), the combinations not present in the original training data (see \cref{fig:waterbirds-dataset})}
    \label{fig:waterbirds-all-tokens}
\end{figure*}

\begin{figure*}[t]
    \includegraphics[width=\textwidth]{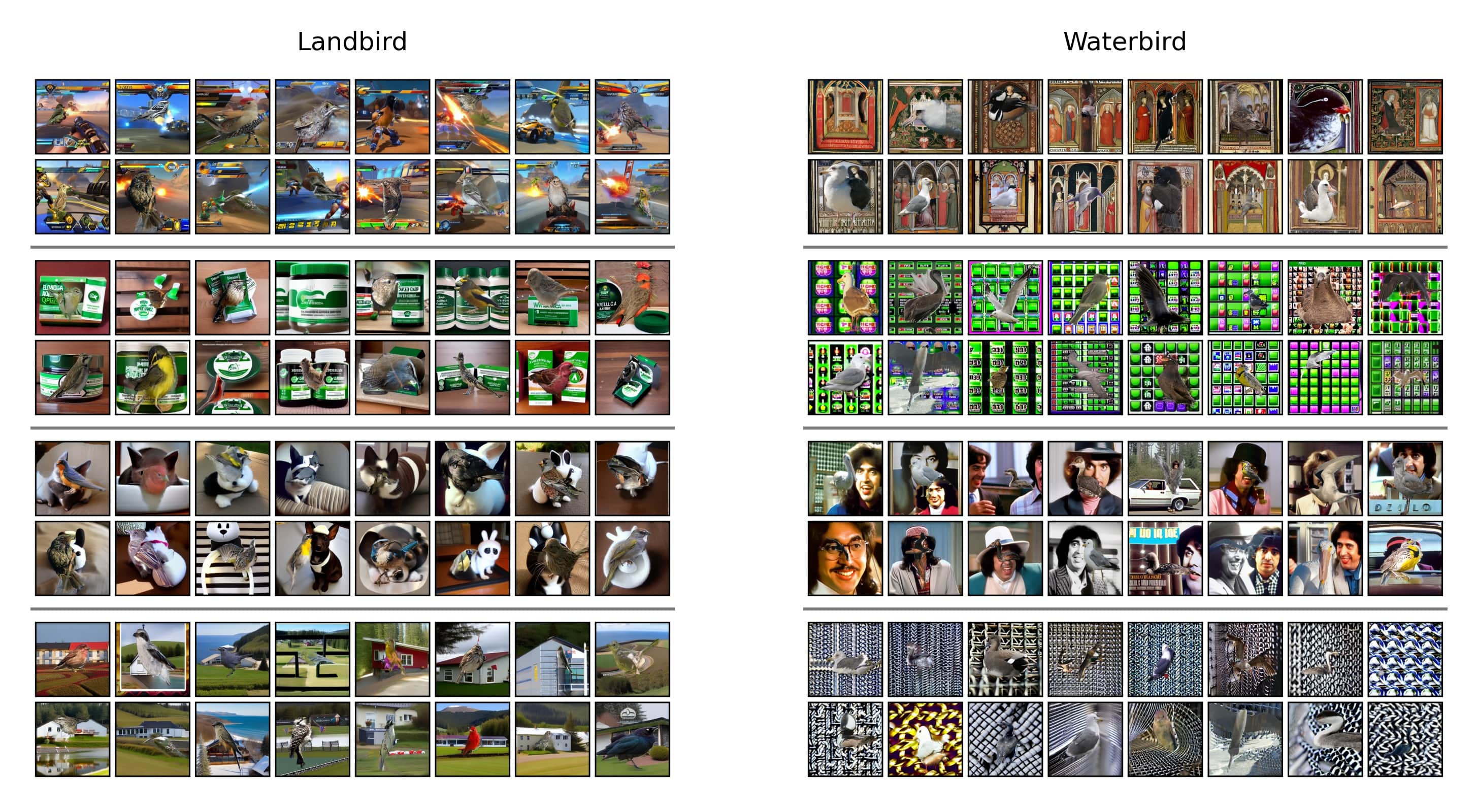}
    \caption{
    \captiontitle{Generation examples for \seabornorange{\aplong}}. 
    Each column shows generations for the corresponding class.
    Each row in a column (separated by gray lines) shows generations for one found token.
    There are eight tokens in total, four for each class.
    While \seabornorange{\ap} finds tokens that fool the model, the generated images are different from the target distribution (land or water background).
    % thus, not useful to adapt the model to it.
    }
    \label{fig:waterbirds-all-adv-tokens}

\end{figure*}

In \cref{fig:waterbirds-all-tokens} and \cref{fig:waterbirds-all-adv-tokens}, we show a few generations using all 8 prompts used in the Waterbirds experiments for \gaplong~and \aplong, respectively.

% \begin{figure}[t]
%     \centering
%     \includegraphics[width=0.9\columnwidth]{figures/appendix/waterbirds_quant_standard_augs.pdf}
%     \caption{
%     \captiontitle{Quantitative results for Waterbirds with additional ``standard'' augmentation baselines.}
%     This plot is similar to Fig.~3-left.
%     Here, we provide additional results for training with RandAugment~\cite{cubuk2020randaugment} and CutMix~\cite{yun2019cutmix} augmentations applied to the original training data.
%     We train a classification model on the original spuriously correlated dataset with the varying number of extra data points generated using different types of prompts.
%     We measure the accuracy on a balanced set where waterbirds and landbirds appear on both land and water. 
%     We run each experiment with three seeds and report the mean and standard deviation.
%     }
%     \label{fig:waterbirds-standard-augs}
% \end{figure}

\subsection{iWildCam}
\label{app:iwildcam}

\subsubsection{Dataset details.}
\begin{figure}[t]
    \centering
    \includegraphics[width=0.48\columnwidth]{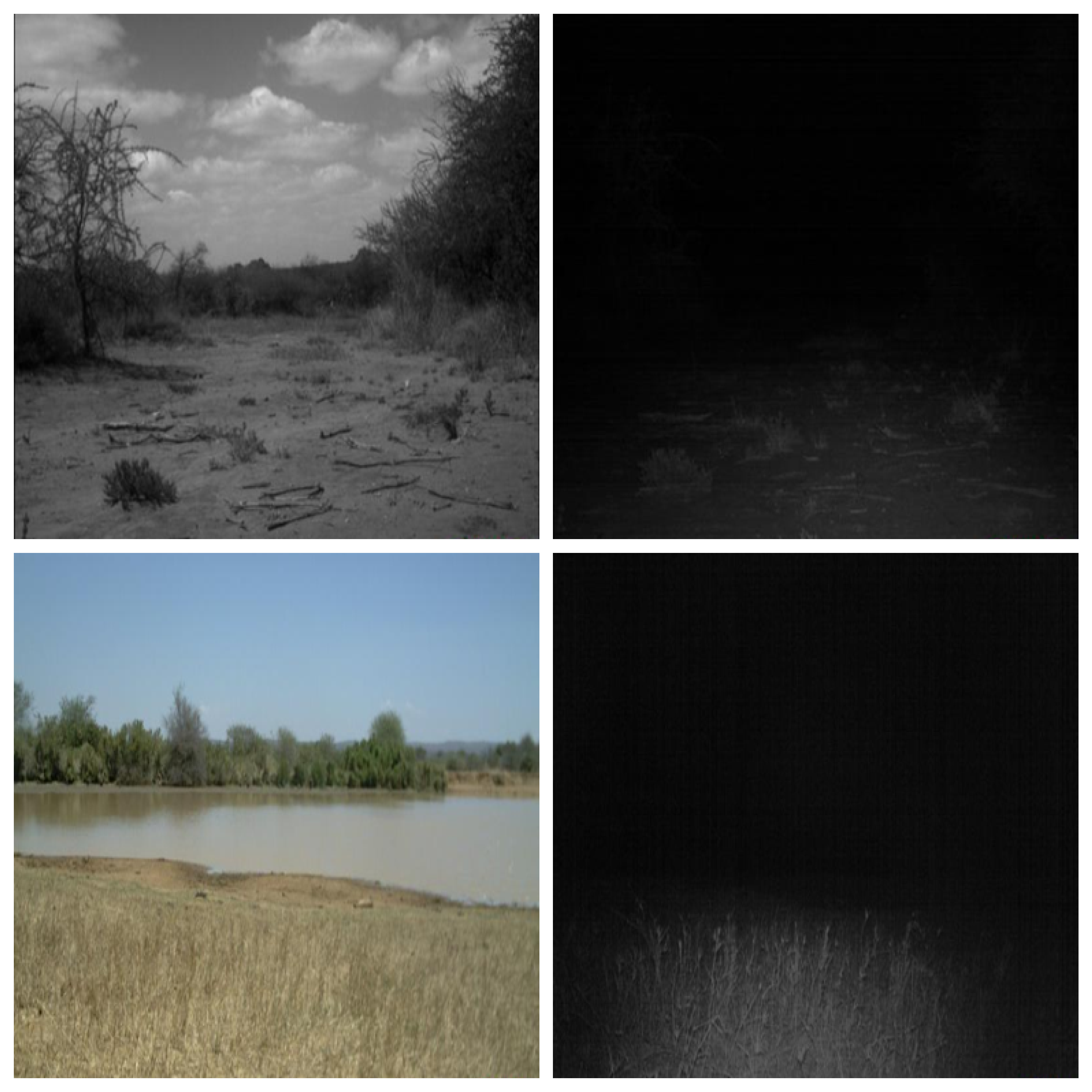}
    \caption{\textbf{iWildCam Test Locations.} Random samples from the four target distributions. First row is \textsc{location=1}, day \& night. Second row is \textsc{location=2}, day \& night.}
    \label{fig:iwild-test-locations}
\end{figure}

The original iWildCam~\cite{beery2021iwildcam} dataset is subsampled to create a 7-way classification task (background, cattle, elephant, impala, zebra, giraffe, dik-dik). The training set has 6,000 images with some classes having as few as 50 images per example. There are 2 test locations that are not in the training or validation set. Additionally, given $h$, the hour at which an image was taken, we define an image to be during \quotes{daytime} if $9 \leq h \leq 17$, and \quotes{nighttime} if $h \leq 5 \lor h \geq 20$. As said in main paper~\cref{sec:classification}, for image CLIP guidance, we separate the target test locations into four groups (\texttt{location=\{1,2\}, time=\{daytime, nighttime\}}).  %Sec.~4.1
We provide visualisation of the test locations (at day \& night) in \cref{fig:iwild-test-locations}. For more details on the iWildCam subset construction, we refer to \cite{dunlap2023diversify} Section 8.3. For inpainting, the object masks are obtained from MegaDetector~\cite{beery_efficient_2019}.

\subsubsection{Alignment collapse solution for iWildCam.}\label{app:iwild-alignment-collapse}
\begin{figure}[t]
    \centering
    \includegraphics[width=0.48\columnwidth]{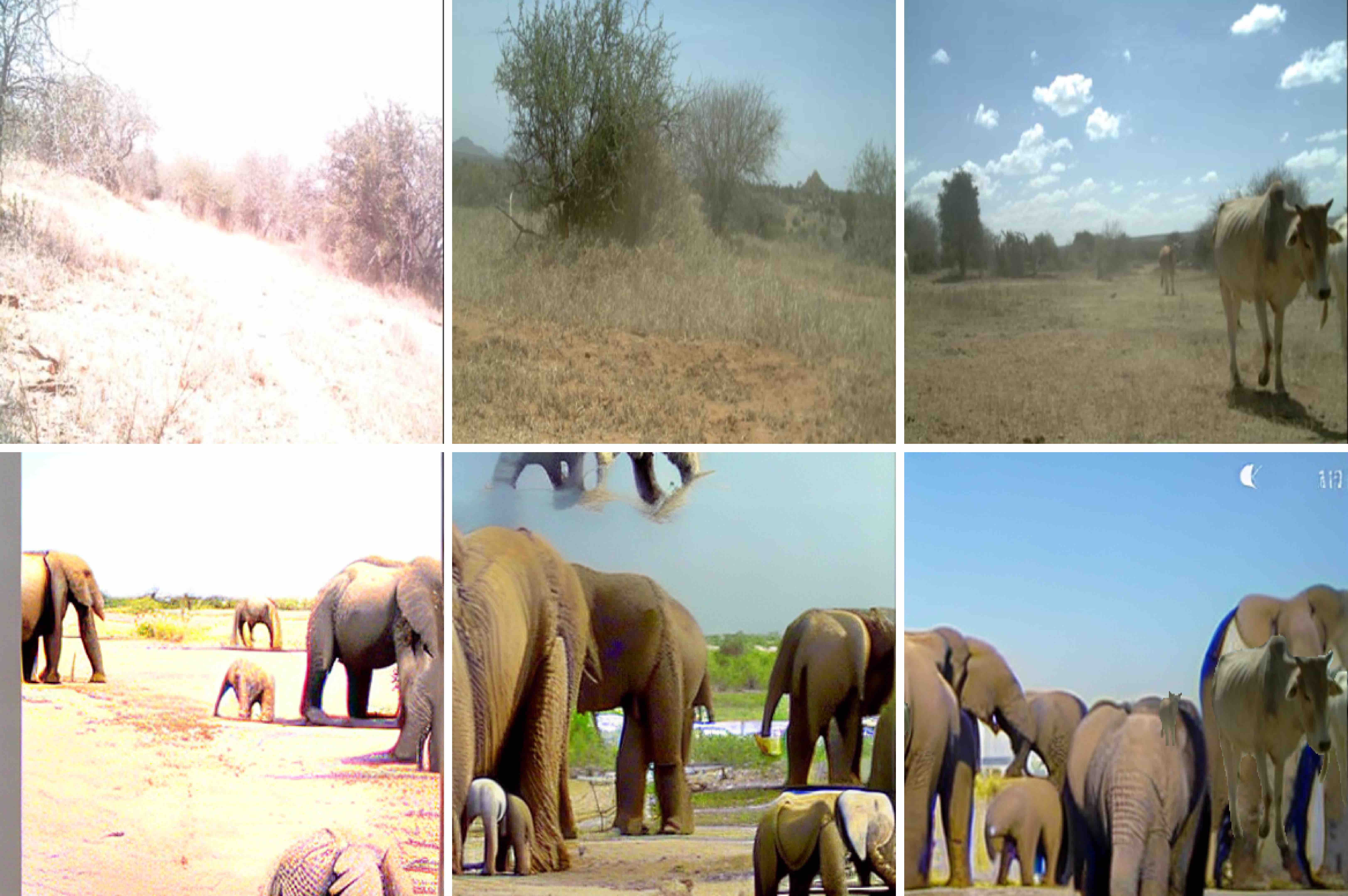}
    \caption{\textbf{Using the negative cross-entropy loss may lead to adversarial animal (e.g., elephants) insertions, destroying the alignment between $\xsample$ and $y$.} First row contains the original training images. The labels are [\textit{background}, \textit{background}, \textit{cattle}]. Second row contains the corresponding generated samples using a guided adversarial prompt, optimized with negative cross-entropy loss as the adversarial loss.}
    \label{fig:iwild-cross-entropy-examples}
\end{figure}

% Sec.~3.1
As mentioned in main paper \cref{sec:prelim}, choosing $\L_\mathrm{adv}$ to be the negative cross entropy loss, i.e. minimizing the probability that the model predicts $y$, may not be the best choice. Indeed, given we use a random sample of 64 images to create our target embedding for the image CLIP guidance, the likelihood that animals were present on these 64 images is very high. This means that the target embedding, although mostly containing the \quotes{location concept}, also partly contains an \quotes{animal concept}. This means that the image CLIP guidance does not explicitly forbid the generation of new animals. Combined with optimizing the negative cross entropy loss, this leads to \textbf{adversarial animal insertions} at generation time, where a new animal of class $\hat{y}$ appears alongside the original animal of class $y$, destroying the $(\xsample,y)$ alignment. In \cref{fig:iwild-cross-entropy-examples}, we provide qualitative examples for this behaviour. To counter this behaviour, we choose  $\L_\mathrm{adv}$ to be the \quotes{entropy} loss, or uncertainty loss. More precisely, this loss is equal to the cross entropy loss where the target label $y$ is replaced by the soft label $\Tilde{y} = [\frac{1}{|\mathcal{Y}|}, \cdots, \frac{1}{|\mathcal{Y}|} ]$, the uniform distribution over all classes. This loss explicitly encourages generations that either (1) do not contain new animals (2) contain new animals that are not accounted for in the label space $\mathcal{Y}$.

\subsubsection{Implementation details}
\label{app:iwildcam-implementation-dets}

\paragraph{Adversarial optimization.}
We describe here the parameters and settings used for optimization. If not precised, the same parameters were used for \aplong~and \gaplong. As said in main paper \cref{sec:classification}, we optimize 4 prompts. Each prompt is composed of 10 placeholder tokens. %The token embeddings at the start of optimization are randomly sampled from $\mathcal{N}(\mu_{emb},\sigma_{emb})$ where $\mu_{emb}$ and $\sigma_{emb}$ is the mean and standard deviation of all embeddings in the vocabulary. 

For optimization, we use a constant learning rate of 0.001, and a batch size of 8.

We use the \quotes{entropy} loss, described previously. For adversarial prompts, we train for 2000 steps. For guided adversarial prompts, we use CLIP guidance coefficient with $\lambda_i = 10$ and and $\lambda_\t{t} = 0$ (i.e., no text guidance). We train for a total of 10000 gradient steps. However, we don't optimize the adversarial loss for the first 2000 steps to allow the prompt to first converge to the target distribution region. One run takes about 3 hours, using one A100 GPU with 80GB of RAM, on our internal cluster.

For adversarial prompts, to generate 4 different prompts, we simply change the seed. For guided (adversarial) prompts, each prompt is w.r.t a new location \& time of the day of the test distribution.

\paragraph{Training data.}
The generation settings are the same as the ones used during adversarial optimization. For each target domain-guided adversarial prompt, (i.e. location \& time of the day), the source images (used to condition the generation with an object mask and through SDEdit~\cite{meng2021sdedit}) are only images that match the time of the day of the target domain used during generation. Furthermore, for each prompt, we only generate one image per source image.

For ALIA, for each prompt, we generate one image per source image, from the whole training dataset. For the generation settings, given we use a slightly different generation process (inpainting) compared to their original implementation, we search ALIA's best-performing generation parameters (according to validation accuracy) over SDEDit strength [0.4, 0.5, 0.8] and guidance scale [5.0, 7.5].  We found the best-performing parameters for ALIA to be the same as the one reported by ALIA in their Github\footnote{https://github.com/lisadunlap/ALIA} i.e. SDEdit strength of 0.5 and guidance of 7.5. 

\paragraph{Finetuning.}
The learning rate scheduler is a cosine scheduler, updated every epoch. The batch size is 128.

Our iWildCam pretrained model is taken from ALIA checkpoints\footnote{https://api.wandb.ai/files/clipinvariance/ALIA-iWildCamMini/brr7b3ks/checkpoint/ckpt-iWildCamMini-randaug-filtered-resnet50-0-0.001-0.0001/best.pth}. ALIA trains the model from ``scratch'' (i.e. the model has never seen iWildCam data), for 100 epochs, on the combination of real + generated data. For our method, given we optimize the prompts based on a finetuned model feedback, it may not make as much sense to train the model from "scratch". Thus, we also introduce the variant where the iWildCam pretrained model is finetuned on the combination of real + generated data for 20 epochs, where finetuning means that every layer, except the last, is frozen. 

For a fair comparison, both training settings are tested for ALIA and our method. We found that ALIA worked best when training from scratch and our method worked best when using the finetuning setting. 

Finally, in their iWildCam experiment, ALIA fixed the number of extra generated points to be used in combination with real data during training to 2224 images. For the sake of comparison, we adopt the same limit in our experiments, with the added variant where the limit is 556 images, showcasing the data efficiency of our method.

\subsubsection{Additional Qualitative Results.}
\label{app:iwild-additional-qual}
\begin{figure}[t]
    \centering 
    \includegraphics[width=0.9\columnwidth]{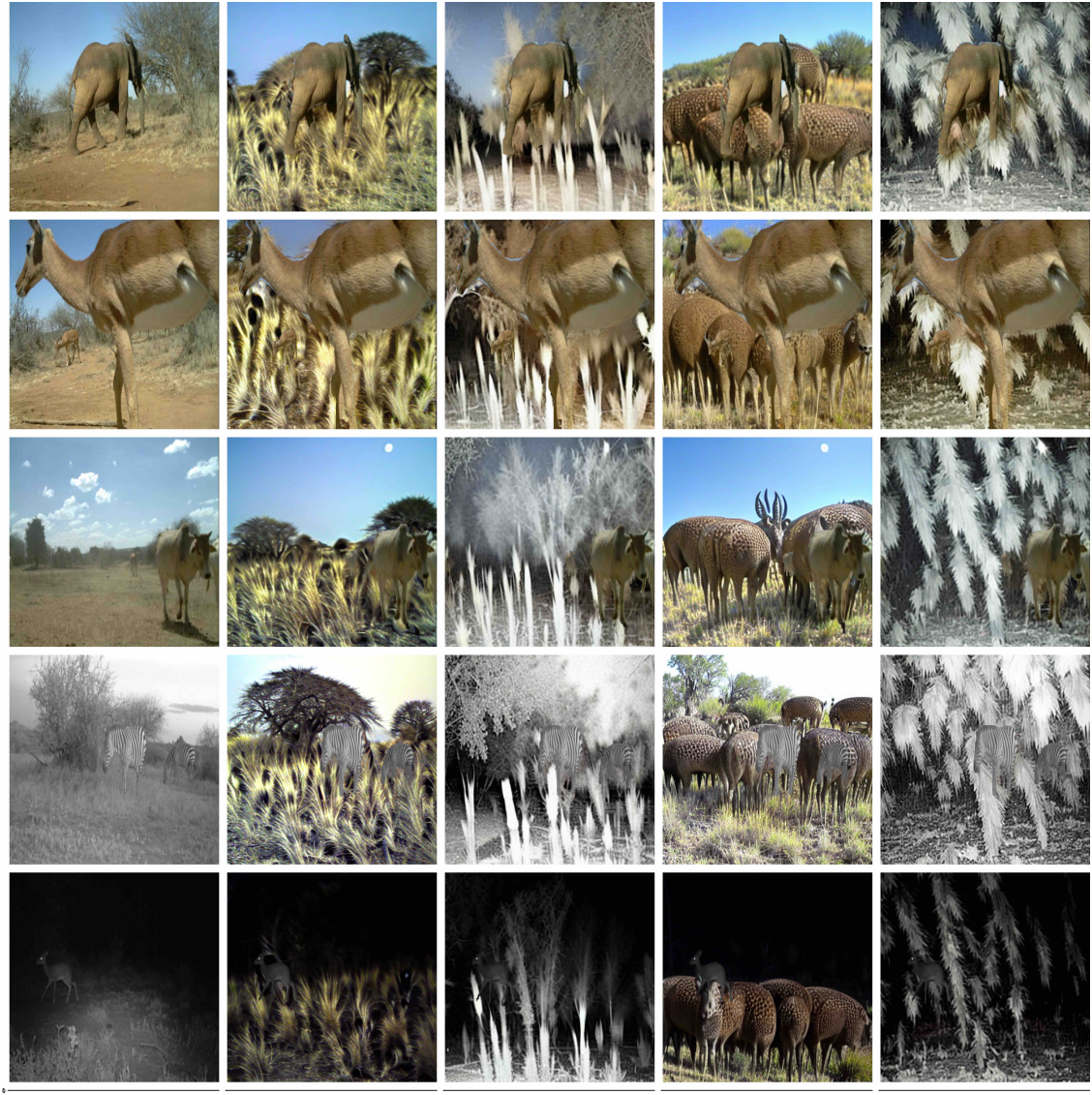}
    \caption{\textbf{Generations with the 4 guided adversarial prompts for iWildCam}. 1st column is the original image. Then from left from right, we have the guided adversarial prompts for \textsc{location 1} during the day, during the night, and then \textsc{location 2} during the day, during the night. SDEdit (strength $0.8$) was used during the
    adversarial optimization and generation.}
    \label{fig:iwild-all-tokens-qual}
\end{figure}

\begin{figure}[t] 
    \centering 
    \includegraphics[width=0.47\textwidth]{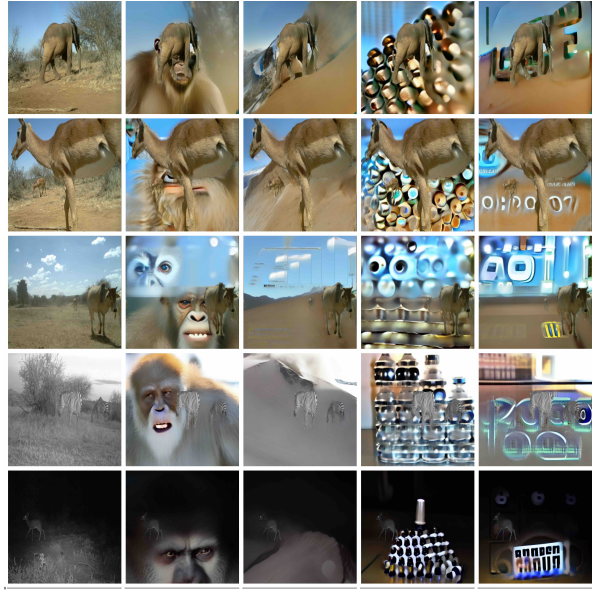}
    \caption{\textbf{Generations with the 4 adversarial prompts for iWildCam}. 1st column is the original image. Then, each column is an adversarial prompt initialized with a different seed. SDEdit (strength $0.8$) was used during the
    adversarial optimization and generation.}
    \label{fig:iwild-all-adv-tokens-qual}     
\end{figure}

In \cref{fig:iwild-all-tokens-qual} and \cref{fig:iwild-all-adv-tokens-qual}, we show a few generations using each of the 4 \gaplong~and \aplong~used in the iWildCam experiments.

% \subsubsection{Additional Quantitative Results}
% \label{app:iwilds-augs}

% In \cref{fig:iwild-quant-appendix}, we provide results with additional baselines.

% \begin{figure}[t] 
% \centering 
% \includegraphics[width=0.97\columnwidth]{figures/appendix/iwild_quant_appendix.pdf}
% \vspace{-1mm}
% \caption{\textbf{Additional quantitative results on iWildCam.} We train a model on
% the combination of the original training data and extra data generated using different types of prompts. We show the average accuracy on two iWildCam test camera trap locations. We run each experiment with three seeds and report the mean and standard deviation. Here, we provide additional results for training with RandAugment~\cite{cubuk2020randaugment} and CutMix~\cite{yun2019cutmix}.}
% \label{fig:iwild-quant-appendix} 
% \end{figure}

\subsubsection{Using ViT model.}

\begin{figure*}[htbp]
    \centering
    \includegraphics[width=0.5\textwidth]{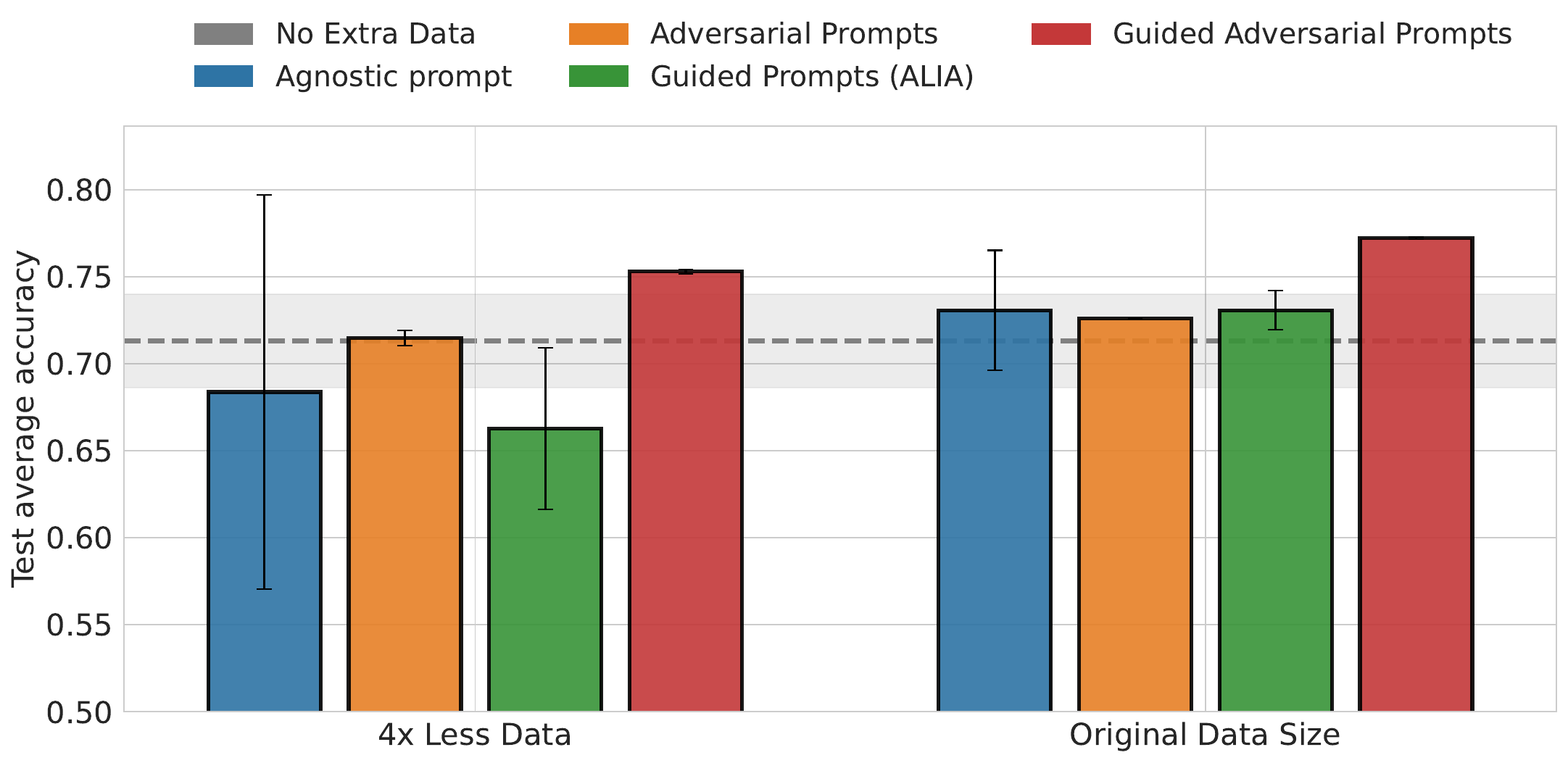}
    \caption{
    We train a ViT-B/16 model on the combination of the original training data and extra data generated using different types of prompts.  
    We show the average accuracy on two iWildCam test camera trap locations.
    We run each experiment with three seeds and report the mean and standard deviation.
} 
    \label{app:iwild_vit_quant}
\end{figure*}

\begin{figure*}[htbp]
    \centering
    \includegraphics[width=\textwidth]{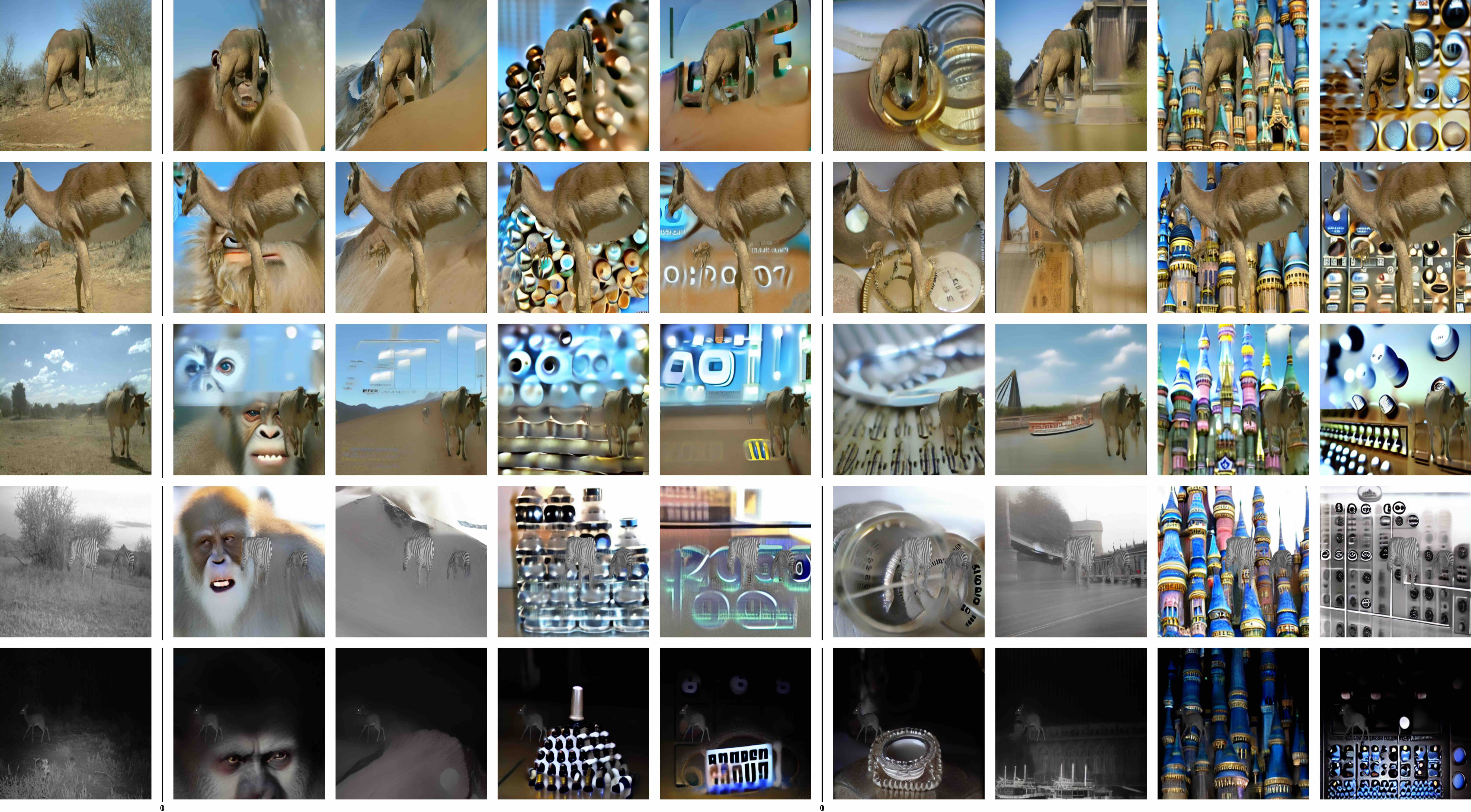}
    \vspace{-5mm}
    \caption{ \textbf{A comparison of generations with adversarial optimization on different models.} First row is original image. Second to fith row are generated using the four ResNet-based adversarial prompts. The four last rows are generated using the four ViT-based adversarial prompts.
} 
    \label{app:iwild_resnet_vit_qual}
\end{figure*}

In \cref{app:iwild_vit_quant}, we repeat the iWildCam experiment from the main paper (\cref{fig:iwild-joint}) with a ViT-B-16~\cite{dosovitskiy2020image} model. Additionally, we provide qualitative results for generations from adversarial prompts optimized with a ViT-B-16 model in \cref{app:iwild_resnet_vit_qual}.

% \begin{figure}[t]
%     \centering
%     \includegraphics[width=\columnwidth]{figures/iwild_model_generalization_quant.pdf}
%     \vspace{-5mm}
%     \caption{
%     We train either a ResNet50 or a ViT-B/16 model on the combination of the original training data and extra data generated using (guided) adversarial prompts, optimized using a pretrained ResNet50 or a ViT-B/16 model.  
%     We show the average accuracy on two iWildCam test camera trap locations.
%     We run each experiment with three seeds and report the mean and standard deviation.
% } 
%     \label{fig:iwild_model_generalization_quant}
%     \vspace{-5mm}
% \end{figure}

\subsection{Image Guidance using Textual Inversion}
\label{sec:ti-guidance} %Sec.~3.3
In addition to the CLIP image guidance introduced in main paper \cref{sec:clip-guidance}, we also explore using Textual Inversion (TI)~\cite{gal2022image} as an image guidance mechanism.
Similar to the CLIP guidance, we use a few images $\{x_j\}$ from the target distribution.
Now, instead of the similarity in the CLIP embedding space, we use the denoising loss between a generated image and one of the target images (see Eq. (2) in~\cite{gal2022image}):
\begin{align}
    \label{eq:ti-guidance}
    \L_\text{TI}(c_w) =& \E_{x_j, z\sim \mathcal{E}(x_j), \epsilon \sim \mathcal{N}(0, I), t\sim U(0, 1)}\\
    &\left[ \|\epsilon - \epsilon_\theta(z_t, t, c_w)\|_2^2 \right],
\end{align}
where $\mathcal{E}$ is the VAE~\cite{kingma_auto-encoding_2022} image encoder and $\epsilon_\theta$ is the denoising UNet model from the Stable Diffusion model~\cite{rombach2022high}, and $x_j$ is sampled randomly from the set of available target images.

We test the TI image guidance on the Waterbirds dataset.
We use the guidance loss from \cref{eq:ti-guidance} with the weight 1000 (we found lower values to result in generations less faithful to the target images) and randomly sample 50 (unlabeled) images from the validation split of the original Waterbirds dataset where both types of birds appear on both types of backgrounds.
We keep other settings the same as for \gap~and \ap.

\begin{figure}
    \centering
    \includegraphics[width=0.5\columnwidth]{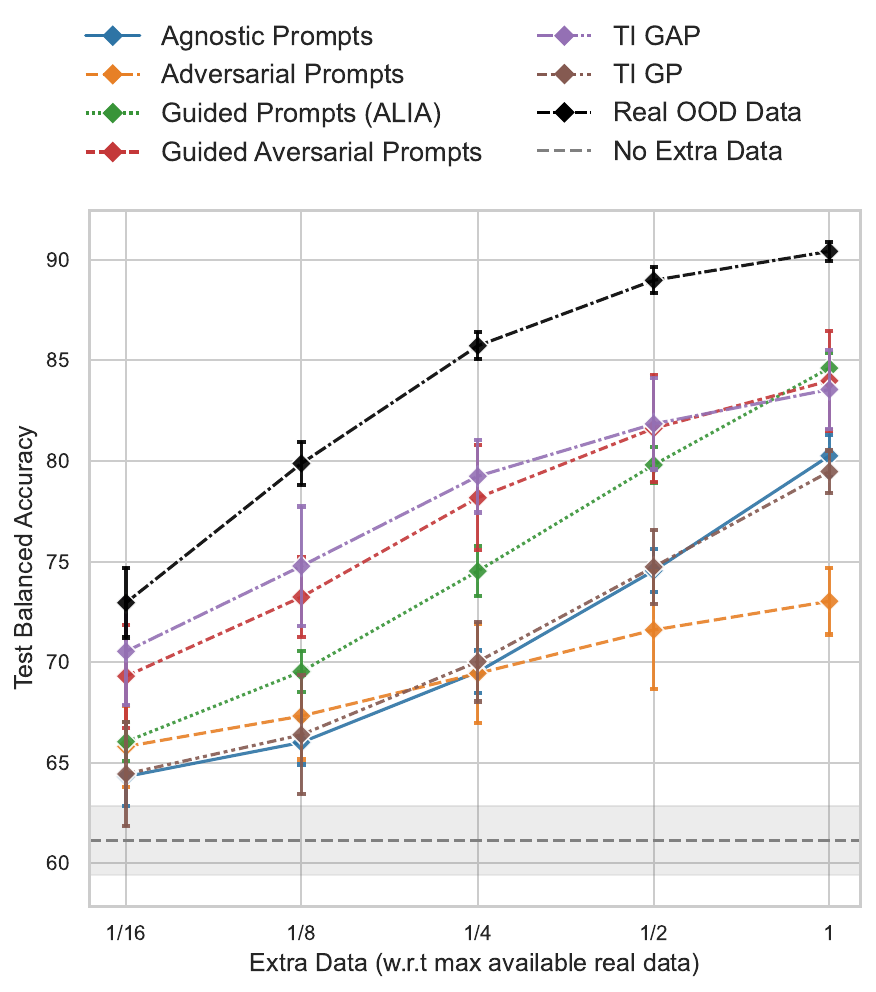}
    \caption{
    \captiontitle{Results of using Textual Inversion image guidance.}
    We use the same plot as in 
    % \cref{fig:waterbirds-standard-augs}.
    \cref{fig:wb-results}.
    We add two additional methods using Textual Inversion image guidance described in \cref{sec:ti-guidance}.
    TI GAP is the \gaplong~with TI image guidance instead of the CLIP guidance.
    TI GP uses only TI guidance to find prompts.
    The TI guidance works on par with the CLIP guidance when used with adversarial optimization (TI GAP).
    However, using only TI guidance (TI GP) results in worse performance than using only text guidance with prompts found by the ALIA method~\cite{dunlap2023diversify}.
    }
    \label{fig:waterbird-ti-gap}
\end{figure}

\cref{fig:waterbird-ti-gap} shows that TI guidance works on par with or better than CLIP guidance on the Waterbirds dataset.
We found, however, that the TI guidance does not result in faithful generations for iWildsCam dataset, and further investigations are needed.

\subsection{Additional Analysis}\label{app:classification-additional-analysis}

% At the \underline{formulation level}, our approach of employing model-informed and target distribution-informed feedback to generate training data is \textbf{general and can be applied to any conditional image generator and supervised model}.

% At the \underline{implementation level}, our approach does require some customization of, e.g., the loss function or conditioning of the generative model.
% We ablated hyperparameters like using l2 or cosine loss for l_clip, cross-entropy or uncertainty for the l_adv loss. We chose the one that gave better results. See Fig x for the results of the former and L239-248 for details on the latter.

% We ablate hyperparameters like \textbf{1)} using $\ell_2$ or cosine loss for $\mathcal{L}_\text{CLIP}$, \textbf{2)} cross-entropy or uncertainty for the $\mathcal{L}_\text{adv}$ loss or \textbf{3)} different ways of incorporating guidance e.g., text or image with CLIP or textual inversion~(T.I)\. We chose the one that gave the best results. 
% See L239-248 for details on \textbf{2} and Appendix 2.6 for T.I.

We ablate hyperparameters like \textbf{1)} using $\ell_2$ or cosine loss for $\mathcal{L}_\text{CLIP}$, \textbf{2)} different ways of incorporating guidance e.g., text or image with CLIP or textual inversion~(T.I). 
% We chose the one that gave the best results. 
\begin{table}
\centering
\footnotesize
\begin{tabular}{l|l|l|l}
\toprule
Loss\textbackslash Guide. & Text & Image & T.I.\\ \midrule
$\ell_2$ & 84.0 & 77.3 & 83.6 \\ 
Cosine sim. & 82.5 & 72.7 & --- \\ \bottomrule
\end{tabular}
\caption{Results from different hyperparameters, namely, loss for CLIP guidance and how guidance is incorporated, on the Waterbird dataset.}
\end{table}

The table shows the accuracy from different combinations of \textbf{1} and \textbf{2} on the Waterbirds dataset. Using $\ell_2$ and text guidance worked best, thus, we used this setting for our results in~\cref{fig:wb-results}. 
T.I. compares generations to the original images in the pixel space, and Image in the CLIP embedding space, resulting in different guidance mechanisms and, hence, performance.

\section{Depth Estimation}\label{app:depth}

\subsection{Depth training details}\label{app:depth-training-details}

\textbf{Adversarial Optimization.} 
The adversarial optimization was done with AdamW~\cite{loshchilov_decoupled_2019}, learning rate of $5.0\times 10^{-4}$, weight decay of $1.0 \times 10^{-3}$, and batch size of 8. The token embeddings at the start of optimization are randomly sampled from $\mathcal{N}(\mu_{emb},\sigma_{emb})$ where $\mu_{emb}$ and $\sigma_{emb}$ is the mean and standard deviation of all embeddings in the vocabulary. We set the early stopping threshold to 0.08 for the UNet model and 1.0 for the DPT model. \textbf{Note that these models were trained with different losses, $\ell_1$ for the former and Midas loss~\cite{eftekhar_omnidata_2021} for the later.} Adversarial optimization is performed with the same loss as was used for training these models. One run takes about 15 mins using one 80GB A100, on our internal cluster. We perform a total of 32 runs, to get 32 \aplong~for the UNet model and 30 runs for the DPT model. As the DPT model was trained on Omnidata, which is a mix of 5 datasets, we have 6 runs for each dataset. Different number of placeholder tokens were also used for each run as suggested in \cref{fig:depth-multi-iter} of the main paper. For the DPT model, we do 1, 8, 16 tokens runs for each dataset and also 3 runs with 32 tokens for each dataset. For the UNet model, 4 runs of 1, 8 and 16 tokens each and 16 runs of 32 tokens were used, to get a total of 32 prompts. We also use a reduced number of denoising steps during optimization i.e., 5, as we found it to be more stable.

\textbf{Guided Adversarial Optimization.} The CLIP guidance coefficient for text and image guidance is set to 1 and 5 respectively. For image guidance, we randomly sampled 100 images from the target distribution. For text guidance, we used target distribution's name in the prompt, e.g., ``fog'' for the fog corruption from CC.

\textbf{Random prompts.}  In our proposed method, 
we optimize for $n$ embedding vectors, resulting in a prompt, $c$. Thus, to match this setting, from a Gaussian distribution fitted on the embeddings from the vocabulary, we sample $n$ random embeddings to create a random prompt to be used in the data generation.

\textbf{Generation.} 
%We pre-computed the images to be used for fine-tuning. 
Generation is performed with the DDIM~\cite{song2020denoising} scheduler and 15 sampling steps. We generate 80k images for the UNet model and 60k images for the DPT model for fine-tuning. For the GP runs with SDEdit, we used strength 0.6, for the GAP runs, strength 0.9. See 

During optimization, we use only 5 denoising steps, as it is more stable.

\textbf{Fine-tuning.} For fine-tuning, we optimize the UNet model with AMSGrad~\cite{reddi2019convergence} with a learning rate of $5.0\times 10^{-4}$, weight decay of $2.0\times 10^{-6}$  and batch size 128. For the DPT model, a learning rate of $1.0\times 10^{-5}$, weight decay of $2.0\times 10^{-6}$ and batch size 32.

\subsection{Additional Quantitative Results}
\label{app:depth-iwild-baselines}
\subsubsection{Depth Estimation}
\label{app:depth-augs}
\textbf{Performance of non-SD baselines.} In \cref{app-tab:depth-results}, we show the results for depth estimation for two additional baselines, deep augmentation~\cite{hendrycks2021many} and style augmentation~\cite{geirhos_imagenet-trained_2018} that do not make use of generative models. Deep augmentation distorts a given image by passing it through an image-to-image model e.g., VAE~\cite{kingma_auto-encoding_2022}, while perturbing its representations. Style augmentation involves involves applying style transfer to the original training images. They perform comparably to \aplong.

\begin{table}[h]
\centering
\begin{adjustbox}{width=0.65\textwidth}
\begin{tabular}{l|cccc|cc} \toprule
                             & \multicolumn{4}{c|}{U-Net}               & \multicolumn{2}{c}{DPT}                 \\ \cmidrule{2-7}
                             & \multicolumn{3}{c}{Taskonomy} & Replica & \multicolumn{2}{c}{Taskonomy} \\ \midrule
Shift                        & Clean     & CC      & 3DCC    & CDS     & CC            & 3DCC              \\ \midrule
Control (No extra data)                     & 2.35      & 4.93    & 4.79    & 5.38    & 3.76          & 3.42              \\ \midrule
\seabornblue{Agnostic Prompts}             & 2.47      & 5.03    & 4.17    & 5.30    & 4.06          & 3.58              \\
\seabornblue{Agnostic Prompts} (Random)    & 2.38      & 4.96    & 4.11    & 5.14    & 3.88          & 3.51              \\
\seabornorange{\aplong}          & 2.49      & 4.36    & 4.02    & 5.12    & 3.40          & 3.28              \\
\seabornorange{\aplong} (SDEdit) & 2.59      & 4.20    & 3.88    & 4.96    & 3.35             & 3.25                    \\ \midrule\midrule
Deep Augmentation & 2.42 & 4.24 & 3.70 & 5.01    &    2.83    &      3.70   \\ 
Style Augmentation & 2.42 & 4.15 & 3.85 & 5.16   &    2.80    &     	3.10   \\ \bottomrule
\end{tabular}
\end{adjustbox}
\caption{\captiontitle{Additional quantitative results on depth estimation.} $\ell_1$ errors on the depth prediction task for a pre-trained U-Net and DPT model. (Lower is better. UNet losses are multiplied by 100 and DPT losses by 10, for readability. Note that the two models were trained with different losses, thus their numbers are not comparable to each other.). We evaluate on distribution shifts from Common Corruptions~\cite{hendrycks2019benchmarking} (CC), 3D Common Corruptions~\cite{kar_3d_2022} (3DCC) and cross-datasets~(CDS), Replica~\cite{straub_replica_2019}. The results from CC and 3DCC are averaged over all distortions and severity levels on Taskonomy. Our method is able to generate training data that can improve results over the baselines on several distribution shifts. Generations with \seabornorange{\ap} (SDEdit) gives better results than \seabornorange{\ap} under distribution shifts. Thus, also conditioning on the original image seems to be helpful for these shifts. For the DPT model, the trends are similar, \seabornorange{\ap} performs better than the baselines. Deep augmentation and style augmentation do not make use of a generative model for generating extra data. They perform comparably to \seabornorange{\ap}.}
\label{app-tab:depth-results}
\end{table}

\subsubsection{Classification with iWildCam}
\label{app:iwild-dg}

% GroupDRO uses group information e.g., for the Waterbirds dataset, whether the bird is on water or land, to minimize the loss of the worse-case group. In their training data, 95\% of waterbirds are on water and 5\% are on land (worst-case group). However, we following the experimental setup of ALIA where the bird is perfectly correlated with the background i.e., all waterbirds are on water. Thus, GroupDRO is not relevant for our experimental setting.

{We show experimental results for several domain generalization baselines on iWildCam. The results for the synthetic data generation methods are based on the low-data regime, i.e., where we generate 4X less data. Our proposed methods, both AP and GAP, outperform these domain generalizaton baselines.}

\begin{table}[h]
\centering
\begin{adjustbox}{width=0.98\textwidth}
\begin{tabular}{l|lllllll|llll} \toprule
\textbf{Method}   & ResNet50 & \begin{tabular}[c]{@{}l@{}}GroupDRO\\ \citep{sagawa2019distributionally}\end{tabular} & \begin{tabular}[c]{@{}l@{}}ADA\\\citep{volpi2018generalizing}\end{tabular} & \begin{tabular}[c]{@{}l@{}}SagNet\\\citep{nam2021reducing}\end{tabular} & \begin{tabular}[c]{@{}l@{}}L2D\\\citep{wang2021learning}\end{tabular} & \begin{tabular}[c]{@{}l@{}}IRM\\\citep{arjovsky2019invariant}\end{tabular} & \begin{tabular}[c]{@{}l@{}}CausIRL\\\citep{chevalley2022invariant}\end{tabular} & 
\begin{tabular}[c]{@{}l@{}}\seabornblue{Agnostic}\\ \seabornblue{Prompts}\end{tabular} & \seabornorange{\ap}   & \seaborngreen{\gp}   & \seabornred{\gap}  \\ \midrule
\textbf{Accuracy} & 67.5     & 70.8     & 61.0        & 64.5           & 77.4        & 55.4        & 69.6            & 69.8             & 79.3 & 71.2 & 81.2 \\ \bottomrule
\end{tabular}
\end{adjustbox}
\caption{\textbf{Additional quantitative results on iWildCam.} We compare our methods against several domain generalization baselines. \seaborngreen{\gp} and \seabornred{\gap} 
 still outperforms.}
\end{table}

\subsubsection{Training with additional generated data}
\label{app:add-gen-data}

{We experimented with training on more generated data for Waterbirds and Depth estimation. 
We found that the trends are dataset specific. For Waterbirds, GAP underperforms GP with 2X the amount of extra data (relative to the original training data, see ~\cref{fig:wb-results} for the results for up to 1X extra data). However, for depth estimation GAP outperforms GP (see ~\cref{fig:depth-gen-comparison} for the results up to $10^4$ extra data).}

\begin{table}[h]
\centering
\begin{adjustbox}{width=0.6\textwidth}
\begin{tabular}{l|lllllll|llll} \toprule
Dataset    & Waterbirds (Acc.↑) &      & Depth ($\ell_1$ Err.$\times100$ ↓) & \\ \midrule
Extra data & 1X  & 2X   &  $10^4$ & $10^5$ \\ \midrule
\seaborngreen{\gp}      & 84.6               & 89.2 & 3.79 & 3.7\\
\seabornred{\gap}        & 84                 & 87.2 & 3.47 & 3.3 
\\ \bottomrule
\end{tabular}
\end{adjustbox}
\caption{\textbf{Training with additional data.} We extend the results in~\cref{fig:wb-results} and~\cref{fig:depth-gen-comparison} to show the performance of our method with more generated data.}
\end{table}

\subsection{Additional Qualitative Results}
\label{app:depth-additional-qual}

\begin{figure*}
    \centering
    \includegraphics[width=\textwidth]{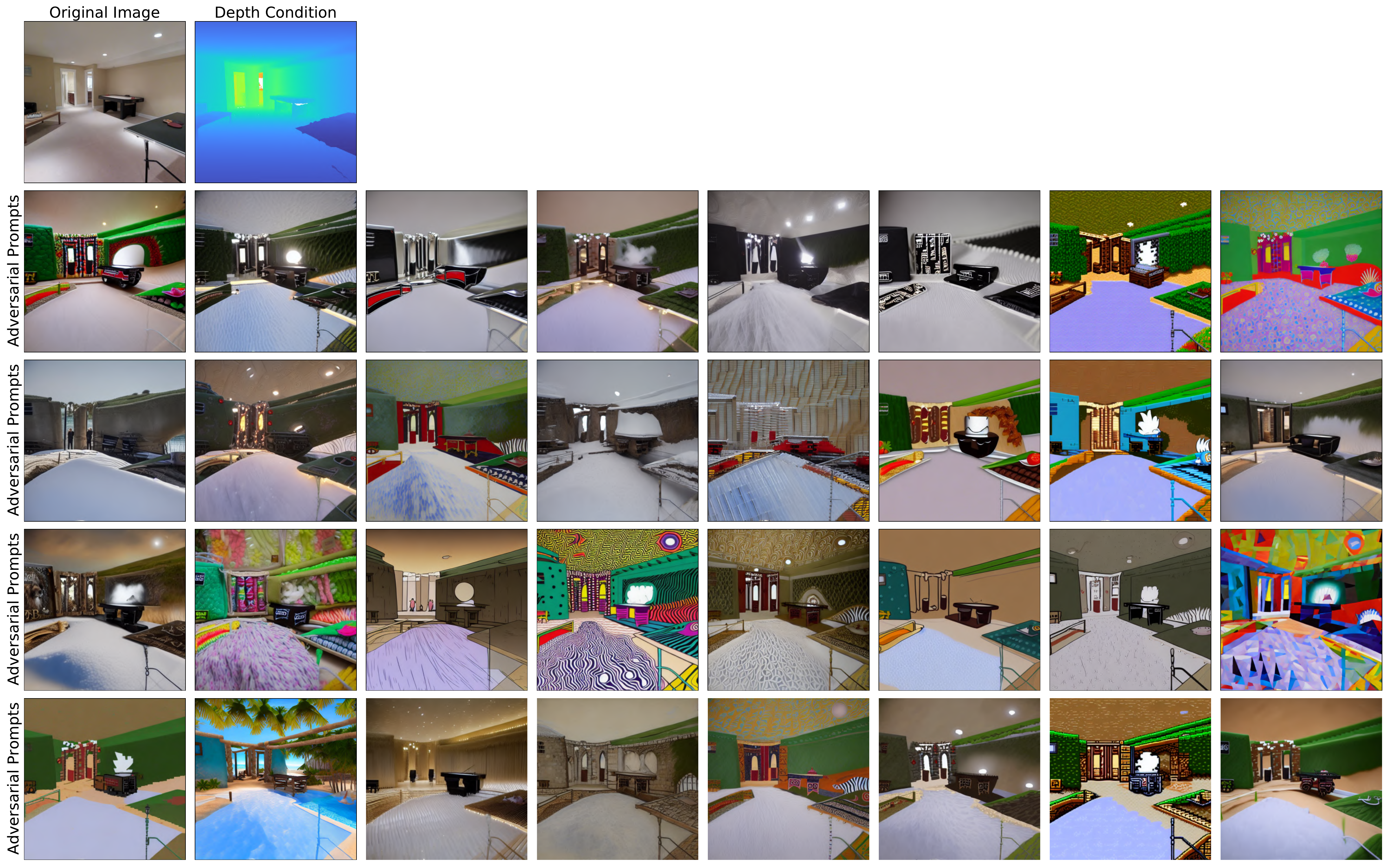}
    \caption{Generations from all \aplong~with the UNet model as the base model. }
    \label{fig:depth-all-ap-prompts-unet}
\end{figure*}

\begin{figure*}
    \centering
    \includegraphics[width=\textwidth]{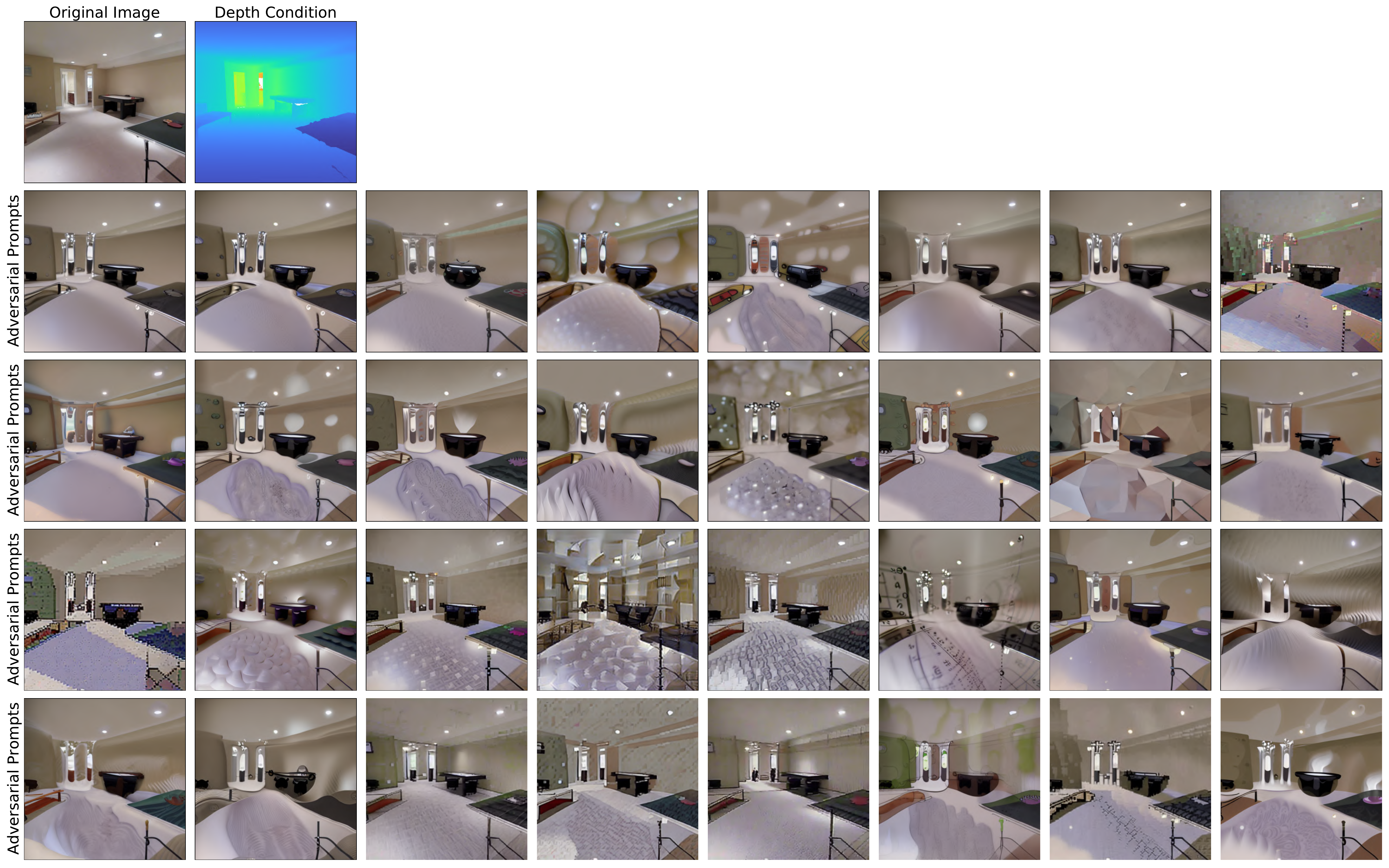}
    \caption{Generations from all \aplong~with the UNet model as the base model. SDEdit (strength 0.6) was used during the adversarial optimization and generation, thus, the generations look similar to the original image. Zoom in to see the different perturbations generated.}
    \label{fig:depth-all-ap-prompts-unet-sdedit}
\end{figure*}

\begin{figure*}
    \centering
    \includegraphics[width=\textwidth]{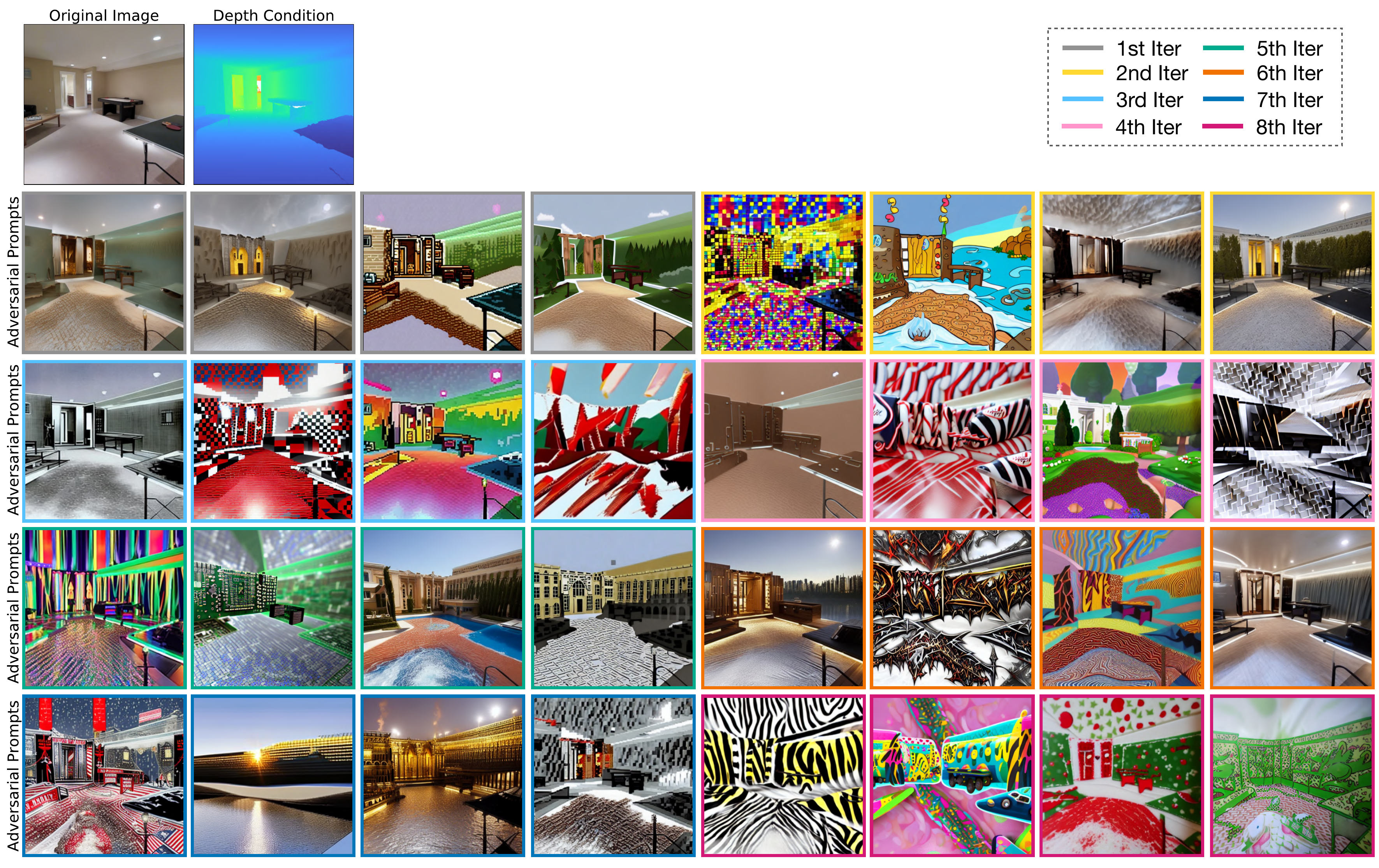}
    \caption{Generations from all \aplong~with the UNet model as the base model for the multi-iteration case i.e., multiple iterations of adversarial optimization, generation and fine-tuning. The colored borders denote the iteration number. Note that we set the early stopping threshold to be 0.1 for the first 3 iterations and 0.08 for the other iterations. We optimized for 4 prompts for each iteration, with an increasing number of tokens for each prompt.}
    % See the main paper \cref{sec:additional-analysis} for details.}
    \label{fig:depth-all-ap-prompts-unet-multi}
\end{figure*}

\begin{figure*}
    \centering
    \includegraphics[width=\textwidth]{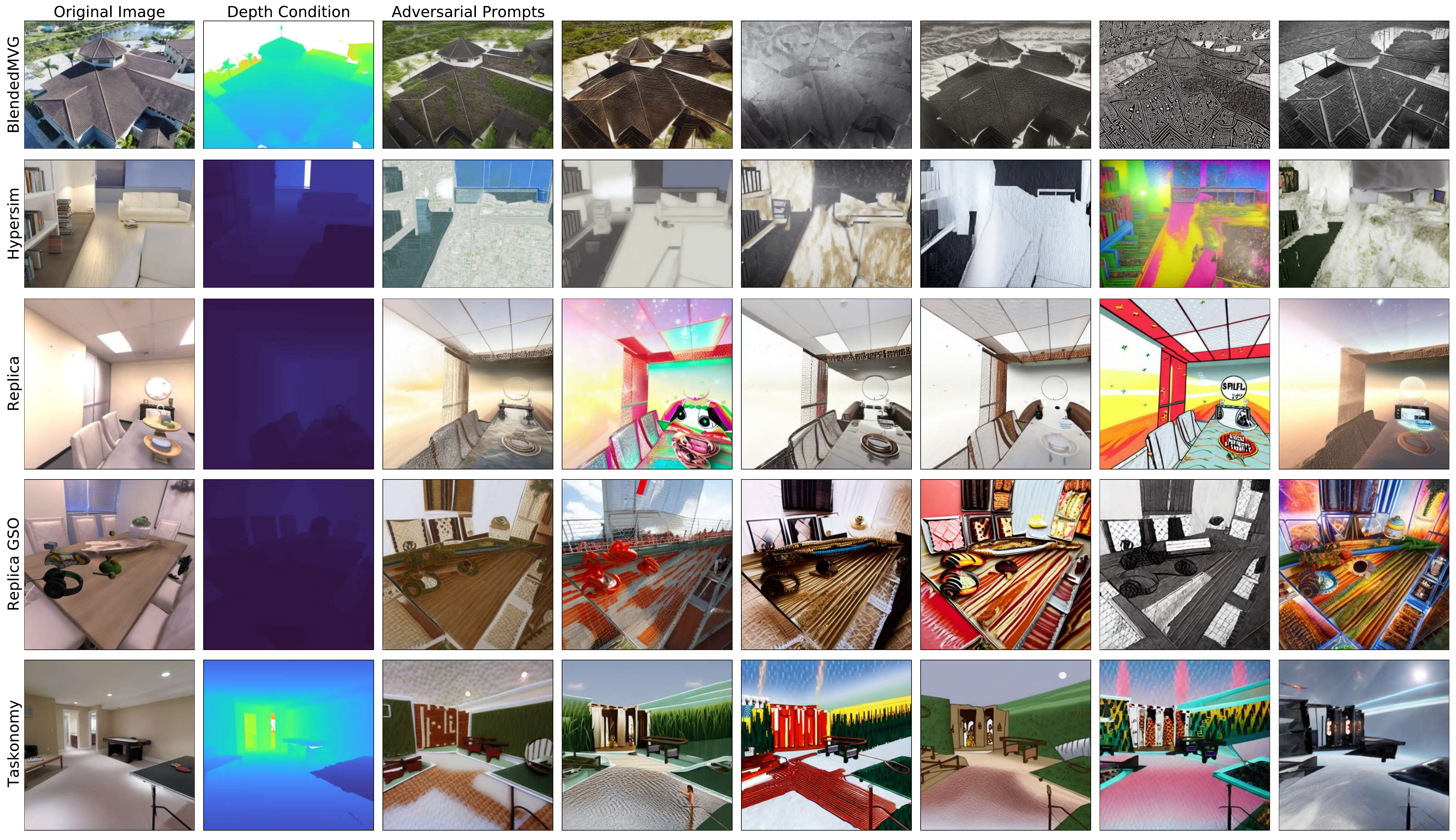}
    \caption{Generations from all \aplong~with the DPT model as the base model. The model was trained on Omnidata which consists of 5 datasets and we optimized for 6 \aplong~for each data. Each row shows the generation from the 6 different prompts for that dataset.}
    \label{fig:depth-all-ap-prompts-dpt}
\end{figure*}

\begin{figure*}
    \centering
    \includegraphics[width=\textwidth]{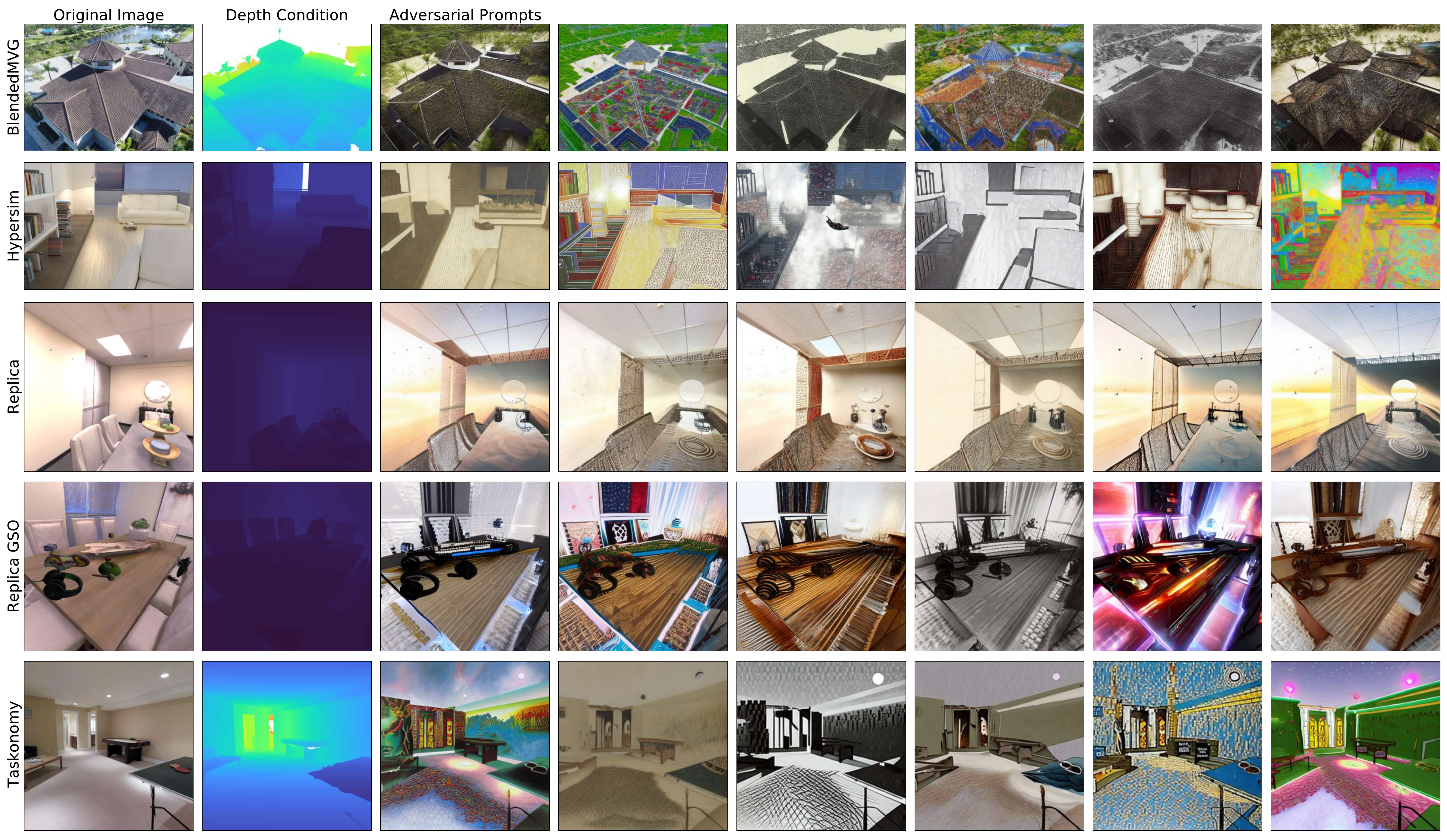}
    \caption{Generations from all \aplong~with a DPT model, also trained on Omnidata. However, this model was also trained with CC and 3DCC augmentations and consistency constraints.}
    \label{fig:depth-all-ap-prompts-dpt-augmented}
\end{figure*}

\textbf{Generations from all adversarial prompts \& comparison of generations from different models.} 
We show the generations from all \aplong~from the UNet model, without SDEDit (\cref{fig:depth-all-ap-prompts-unet}), with SDEdit (\cref{fig:depth-all-ap-prompts-unet-sdedit}), and multi-iteration (\cref{fig:depth-all-ap-prompts-unet-multi}). Additionally, we provide the generations from two DPT models, allowing us to assess the difference the model feedback has on generations. The first DPT model was only trained on Omnidata (\cref{fig:depth-all-ap-prompts-dpt}) and second was trained on Omnidata with augmentations from CC and 3DCC and with consistency constrains~\cite{zamir_robust_2020} (\cref{fig:depth-all-ap-prompts-dpt-augmented}). The quantitative results in the paper were reported only on the former DPT model.

There does not seem to be obvious differences in the styles generated between the two DPT models. However, between the \aplong~from the UNet model with and without multi-iteration, the \aplong~from the latter seems to result in much more diverse styles.

\subsection{Additional Analysis}
\label{app:depth-additional-analysis}
\begin{figure}[h]
    \centering
    \includegraphics[width=0.5\columnwidth]{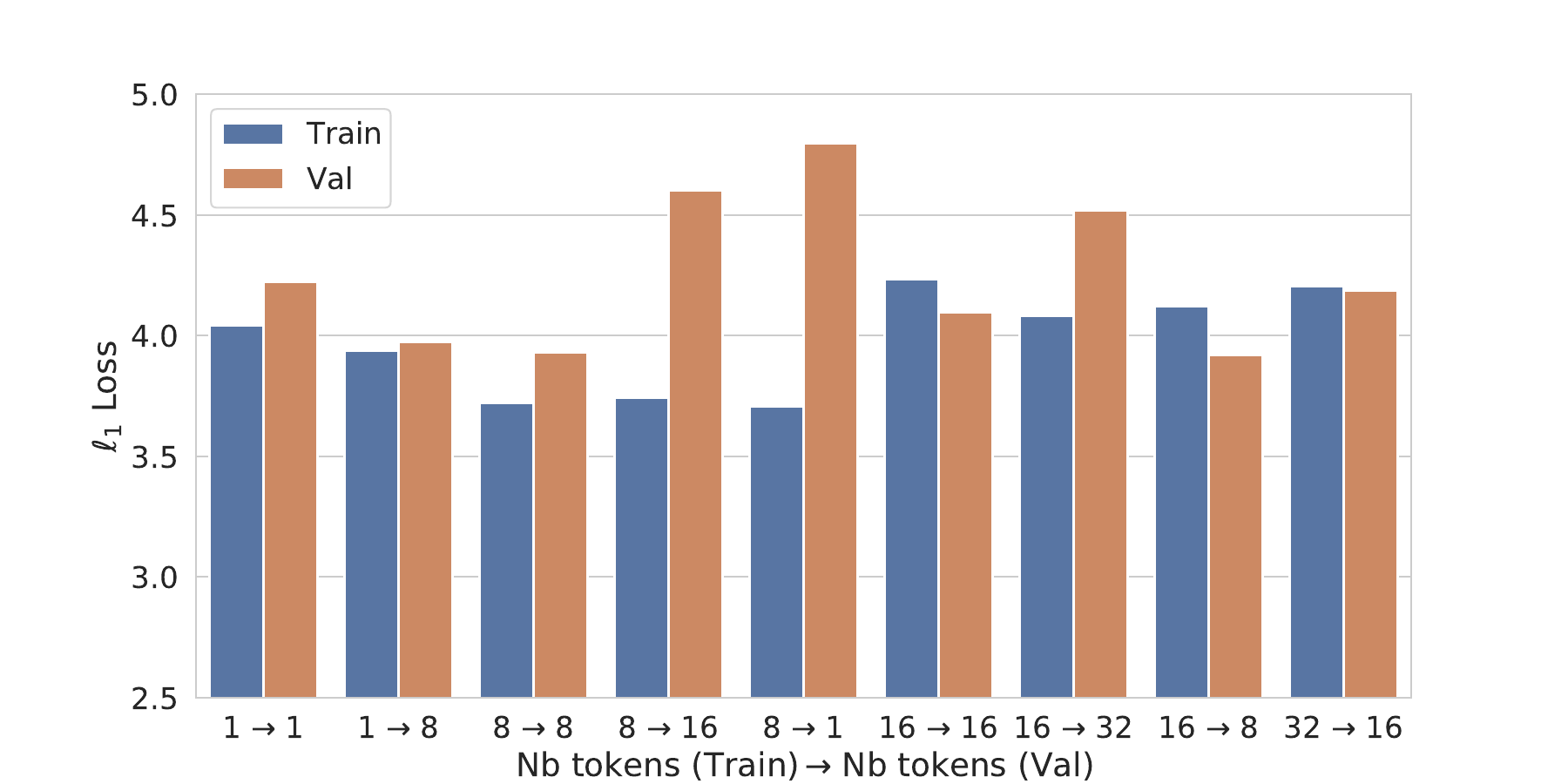}
    \caption{Generalization to generated images from similar or larger number of tokens. This plot shows the performance of the UNet model when trained on \aplong~with $n$ number of tokens and tested on \aplong~with $m$ number of tokens (denoted in the plot as $n\rightarrow m$). With the exception of $1\rightarrow 1$ and $1\rightarrow 8$, training on \aplong~with $n$ tokens and testing on $m, n\neq m$ results in higher loss than training and testing on the same number of tokens.}
    \label{fig:depth-generalization}
\end{figure}

\subsubsection{Running multiple iterations of adversarial optimization vs a single iteration}
\label{app:multi-iter}

Here, we provide additional analysis for the multi-iteration experiments in \cref{fig:depth-multi-iter} in the main paper. 
We optimize for 4 prompts in each iteration and noticed that if the number of placeholder tokens in a given prompt is kept fixed throughout the iterations, the optimization to find new \aplong~becomes more difficult. However, if we increase the number of tokens at each iteration e.g., $1$ token per prompt for 1st, $8$ per prompt for 2nd, etc, we are able to consistently find new \aplong. 
% This seems to imply that the model, once fine-tuned on generated data from a set of prompts with a given number of tokens, is able to generalize to the adversarial generations from new prompts with the same number of tokens. 
% We observed that if we fixed the number of tokens in each iteration, the loss when performing adversarial optimization is not as low in subsequent iterations i.e., the loss in the 2nd or 3rd iteration tends to be higher than that from the 1st iteration. 
Thus, we aim to investigate the generalization of a given model to different \aplong, e.g., is a model more likely to generalize to \aplong~with the same number of tokens. 

% While exploring the multi-iteration setting, we initially set the number of tokens per prompt to be fixed for every iteration. The problem with that setup was that the algorithm can not find Adversarial Prompts starting from then 2nd or 3rd iteration, i.e. the loss does not decrease. That suggests that data generated with the found optimized prompts exhibits common features, and a model fine-tuned on these data will generalize to any other data generated using \ap~with different prompts. 

To perform this analysis, we generated data $D_n, D_m$ using \ap~with $n$ and $m$ tokens per prompt respectively and measured the performance of a model fine-tuned on $D_n$ on $D_m$.

\paragraph{Results for generalization to the same number of tokens.}
In this setting, $n=m$, we use $n = m \in \{1, \ 8, \ 16, \ 32 \}$. For every $n$, we construct $D_n$ and $D_m$ to be generated using $4$ \aplong. We fine-tune the model on $D_n$ and and validate both on $D_n$ and $D_m$ validation sets during the fine-tuning. The results are shown in Fig.~\ref{fig:depth-generalization}. 

% The loss on $D_m$ is similar to the loss on $D_n$, which means that model, trained on $4$ prompts consisting of $n$ tokens will generalize to other $4$ prompts with the same number of tokens. 

\paragraph{Results for generalization to different number of tokens.}
In this setting, $(n, m)$ are $(1, 8), (8, 1), (8, 16), (16, 8), (16, 32), (32, 16) \ $respectively. For every $n$ we fine-tune on $D_n$ and compute the validation loss on $D_n$ and $D_m$. The results are shown in Fig.~\ref{fig:depth-generalization}. As the loss for $n=m$ tends to be more similar then when $n\neq m$, we chose to increase the number of tokens used per prompt in our multi-iteration setting.

% There is no consistent generalization between data generated with prompts consisting of different number of tokens. This suggest that it is profitable to use different \quotes{search spaces}, i.e. defined by different $n$ in this setting, to find different adversarial shifts.

\begin{figure*}[h] 
\centering 
\includegraphics[width=\textwidth]{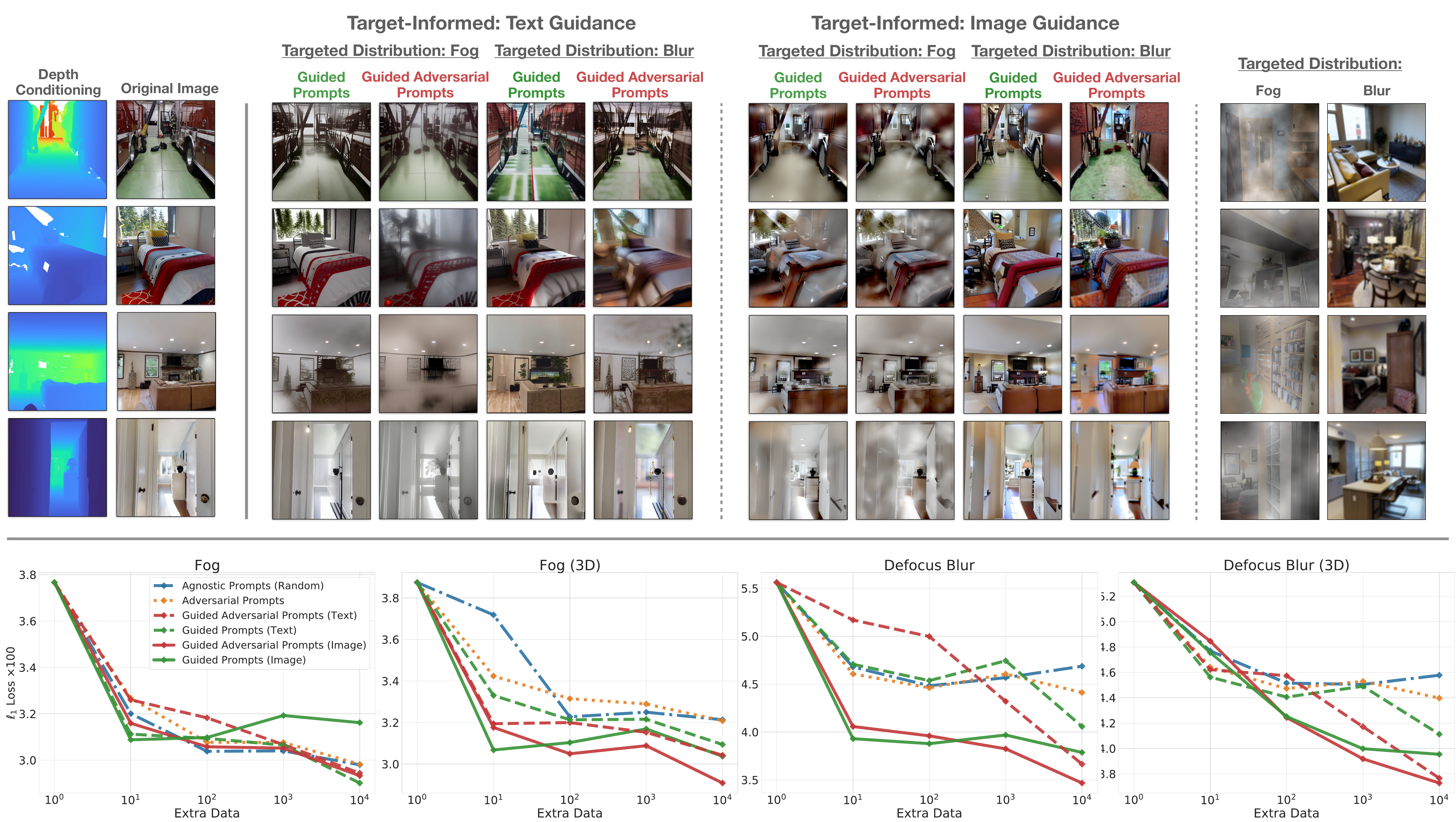}
\caption{\textbf{A comparison of text and image guidance} for two distribution shifts fog and blur. The base model used here is the UNet model.
\textbf{Top:} Generated images from \seaborngreen{Guided Prompts} and \seabornred{\gaplong}. All images were generated with SDEdit, strength 0.9, thus, they tend to look similar to the original image (2nd column).
Using \seaborngreen{\gplong}~alone for either text or image guidance results in generations with a \textit{mild} fog or blur. With \seabornred{\gaplong}, we get generations with \textit{more severe} fog or blur. 
For image guidance, we sampled random (unlabelled) images with blur and fog from the Common Corruptions evaluation set. See the last two columns for some sample images.
\textbf{Bottom:} The plots show the results from fine-tuning with text and image guidance, evaluated on fog and blur for CC and 3DCC. Note that the image guidance here uses (unlabelled) samples from the same target distribution that it was sampled on. In all cases, \seabornred{\gaplong}~outperforms \seaborngreen{Guided Prompts} with large enough extra data. 
% A comparison of the generated data for the baselines and our method. Generations with \seabornorange{\aplong}~results in \textit{diverse styles} that are \textit{distinct from the original training data} (see fifth column). Adversarial optimization and generation with SDEdit results in generations shown in the fourth column. As these generations are conditioned on the original image, they look more similar to them.
% The last 4 columns show the generations with text guidance for target shifts \textit{fog} and \textit{blur}. Using \seaborngreen{\gplong}~alone, in this case ``fog'' or ``blur'', results in generations with a mild fog or blur. Performing \seabornred{\gaplong}~results in generations with more severe fog or blur. 
}
\label{app-fig:depth-gen-comparison} 
\end{figure*}

% \begin{figure*}[h]
%     \centering
%     \includegraphics[width=1.95\columnwidth]{figures/sup-depth-text-vs-img-guide-long.pdf}
%     \caption{\ty{update}}
%     \label{fig:depth-guide-results}
% \end{figure*}
% \cref{fig:depth-guide-results} shows the corresponding quantitative results from fine-tuning on the generated data from both text and image Guided Prompts and \gaplong, on the fog and blur corruptions from CC and 3DCC.
\subsubsection{CLIP image vs text guidance.} 
\label{app:clip-img-vs-text}

In \cref{app-fig:depth-gen-comparison}, we compare the qualitative (top) and quantitative (bottom) differences in generations from text guidance and image guidance on defocus blur and fog.  Note that image guidance uses sample (unlabelled) images from the corresponding target distribution that it is evaluated on, i.e., fog samples images from the fog corruption from the  CC benchmark and fog (3D) samples images from the fog corruption of the 3DCC benchmark. If the target distribution name has (3D) appended to it, it is from the CC benchmark, otherwise it is from the 3DCC benchmark.
% for the target distribution (see last two columns), where the blur or fog is applied uniformly over the image.

We observed some differences in the generations with text vs. image guidance (\cref{app-fig:depth-gen-comparison}, top). Text Guided Prompts generates corruptions that are more realistic that image Guided Prompts. For example, fog gets denser further away from the camera or around the floor when text Guided Prompts are used for generations. For image Guided Prompts, as it was guided by the image samples from CC where the corruption is applied uniformly over the image, it learns to also apply a more uniform corruption over the image. 
We also noticed that the diffusion model was not able to generate certain shifts e.g., compression artifacts, we leave further analysis to future work.
% , as the text descriptions e.g., `noise', was too ambiguous.
% We leave further analysis to future work.

Quantitatively, we observed that image guidance tends to perform the best across the target distributions, with large enough extra data (\cref{app-fig:depth-gen-comparison}, bottom). 
%It outperforms both text guidance and image Guided Prompts. 
% Thus, when evaluated on the CC benchmark, image guidance performs better than text guidance.

% This is reflected in the quantitative results, where text guidance (both Guided Prompts and \gaplong) results in better performance on 3DCC than CC.

% Unlike classification, the performance of \gap~for depth is not consistent across all distribution shifts. 
% The diffusion model was not able to generate certain shifts e.g., noise corruptions, as the text descriptions and unlabelled images was too ambiguous e.g., `noise' or having common attributes other than the corruption. 

\subsubsection{Generalization of \aplong~to different models.} We show how adversarial generations from \aplong~found for one model are for another model in \cref{tab:depth-ap-gen}. The generations from \aplong~found for e.g., the UNet model result in the highest loss when evaluated on the UNet model. However, the generations from \aplong~from the DPT model also result in similar loss. Similar trends hold for the DPT model. Thus, \aplong~found for one model are also able to result in high loss for another model.

\begin{table}[h]
\label{tab:app-analysis-feedback-depth}
\centering
\begin{adjustbox}{width=0.45\textwidth}
\begin{tabular}{l|lll} \toprule
AP from\textbackslash Eval on & Original data & UNet & DPT \\ \midrule
UNet & 2.55 & 7.63 & 5.39 \\
DPT & 1.76 & 7.17 & 6.46 \\ \bottomrule
\end{tabular}
\end{adjustbox}
\caption{Evaluation performance of a model on generated images from \aplong~from another model (without fine-tuning). For the UNet model we report the $\ell_1$ loss ($\times 100$ for readability) and the DPT model, the Midas loss ($\times 10$ for readability). The \aplong~attained from performing adversarial optimization on a UNet model and evaluated on the same model result in a loss of 7.63. Generations from the DPT model evaluated on the UNet model result in a loss of 7.17. Thus, the adversarial prompts found for one model seems to also be adversarial for another. We also report the loss on the original images for comparison.}\label{tab:depth-ap-gen}
\end{table}

% \begin{figure}[h]
%     \centering
%     \includegraphics[width=0.9\columnwidth]{figures/multiiter-32.pdf}
%     \caption{\todo{explain the caption; make "left - right" image} Multi-iteration generalization analysis.}
%     \label{fig:multi-it}
% \end{figure}

% \vspace{-10mm}

% \vspace{-5mm}

\section{Broader Impact} \label{app:broader-impact}
We presented a method to control generative models to produce training data for supervised learning models. 
While our method is not particularly poised for negative use, it should be noted that powerful generative models are a general tool and the method has the potential to be used in ways authors did not intent.
In addition, the data they are trained on may incorporate various societal biases or contain samples
gathered in different ways from the internet. 
Furthermore, we will make our code and models publicly available. Thus, allowing for transparent inspection and safeguarding.

% We developed a framework for training general-purpose foundation models that, as demonstrated, can
% be conveniently re-purposed for various tasks a practitioner may be interested in. We also committed
% to open-sourcing our code and models. These actions support the public with the democratization
% of the tools and the possibility of transparent inspection and safeguarding. While our model is not
% particularly poised for negative use compared to the alternatives, it should be noted that powerful
% generative models are a general tool and have the potential to be used in ways authors did not intent.
% In addition, the data they are trained on may incorporate various societal biases or contain samples
% gathered in different ways from the internet. We trained our models on CC12M [19] which is an
% open-sourced dataset and has been curated to some degree (e.g., people’s names are redacted), yet,
% due to the imperfections in this process, we still advise caution when using the models for generative
% purposes.

% \end{document}

\end{document}